\documentclass[final]{cvpr}

\usepackage{times}
\usepackage{epsfig}
\usepackage{graphicx}
\usepackage{amsmath}
\usepackage{amssymb}

\usepackage[table]{xcolor}
\usepackage{accents}
\usepackage{tabulary}
\usepackage{array}
\usepackage{multirow}
\usepackage{caption}
\usepackage{enumitem}
\usepackage{subfig}
\usepackage{algorithm}
\usepackage{listings}
\usepackage{comment}

\usepackage[pagebackref=true,breaklinks=true,colorlinks,bookmarks=false]{hyperref}

\def\cvprPaperID{2909} 
\def\confYear{CVPR 2021}

\newcommand{\varX}{\mathbf{X}}
\newcommand{\varY}{\mathbf{Y}}

\newcommand{\varQ}{\mathbf{Q}}
\newcommand{\varK}{\mathbf{K}}
\newcommand{\varV}{\mathbf{V}}
\newcommand{\varR}{\mathbf{R}}

\newcommand{\real}{\mathbb{R}}

\newcommand{\localFunc}{\boldsymbol{\phi}}

\newcommand{\varF}{\boldsymbol{\mathcal{F}}}
\newcommand{\varG}{\boldsymbol{\mathcal{G}}}
\newcommand{\varH}{\boldsymbol{\mathcal{H}}}

\newcommand{\apbbox}[1]{AP$^\text{bbox}_\text{#1}$}
\newcommand{\apmask}[1]{AP$^\text{mask}_\text{#1}$}
\newcommand{\ap}[1]{AP$_\text{#1}$}

\newlength\savewidth\newcommand\shline{\noalign{\global\savewidth\arrayrulewidth
		\global\arrayrulewidth 1pt}\hline\noalign{\global\arrayrulewidth\savewidth}}
\newcommand{\tablestyle}[2]{\setlength{\tabcolsep}{#1}\renewcommand{\arraystretch}{#2}\centering\footnotesize}

\newcommand{\cgap}[2]{
	\fontsize{6pt}{1em}\selectfont{(${#1}${#2})}
}
\definecolor{Gray}{gray}{0.5}
\newcommand{\demph}[1]{\textcolor{Gray}{#1}}
\definecolor{Highlight}{HTML}{39b54a}  
\newcommand{\cgaphl}[2]{
	\fontsize{6pt}{1em}\selectfont{\textcolor{Highlight}{(${#1}$\textbf{#2})}}
}
\newcommand{\hl}[1]{\textcolor{Highlight}{#1}}

\begin{document}

\title{Involution: Inverting the Inherence of Convolution for Visual Recognition}


\author{Duo Li\textsuperscript{1}\hspace{0.5em} Jie Hu\textsuperscript{2}\hspace{0.5em} Changhu Wang\textsuperscript{2}\hspace{0.5em} Xiangtai Li\textsuperscript{3}\hspace{0.5em} Qi She\textsuperscript{2}\hspace{0.5em} Lei Zhu\textsuperscript{3}\hspace{0.5em} Tong Zhang\textsuperscript{1}\hspace{0.5em} Qifeng Chen\textsuperscript{1}\\
The Hong Kong University of Science and Technology\textsuperscript{1}\quad ByteDance AI Lab\textsuperscript{2}\quad Peking University\textsuperscript{3} \\
{\tt\small duo.li@connect.ust.hk\quad \{hujie.frank, wangchanghu\}@bytedance.com\quad \{tongzhang, cqf\}@ust.hk}
}

\maketitle
\thispagestyle{empty}

\begin{abstract}
   Convolution has been the core ingredient of modern neural networks, triggering the surge of deep learning in vision. In this work, we rethink the inherent principles of standard convolution for vision tasks, specifically spatial-agnostic and channel-specific. Instead, we present a novel atomic operation for deep neural networks by inverting the aforementioned design principles of convolution, coined as involution. We additionally demystify the recent popular self-attention operator and subsume it into our involution family as an over-complicated instantiation. The proposed involution operator could be leveraged as fundamental bricks to build the new generation of neural networks for visual recognition, powering different deep learning models on several prevalent benchmarks, including ImageNet classification, COCO detection and segmentation, together with Cityscapes segmentation. Our involution-based models improve the performance of convolutional baselines using ResNet-50 by up to 1.6\% top-1 accuracy, 2.5\% and 2.4\% bounding box AP, and 4.7\% mean IoU absolutely while compressing the computational cost to 66\%, 65\%, 72\%, and 57\% on the above benchmarks, respectively. Code and pre-trained models for all the tasks are available at \url{https://github.com/d-li14/involution}.
\end{abstract}


\section{Introduction}

Albeit the rapid advance of neural network architectures, convolution remains the building mainstay of deep neural networks. Drawn inspiration from the classical image filtering methodology, convolution kernels enjoy two remarkable properties that contribute to its magnetism and popularity, namely, spatial-agnostic and channel-specific. In the spatial extent, the former property guarantees the efficiency of convolution kernels by reusing them among different locations and pursues translation equivalence~\cite{pmlr-v97-zhang19a}. In the channel domain, a spectrum of convolution kernels is responsible for collecting diverse information encoded in different channels, satisfying the latter property. Furthermore, modern neural networks appreciate the compactness of convolution kernels via restricting their spatial span to no more than $3 \times 3$, since the advent of the seminal VGGNet~\cite{Sumonyan2015Very}.

On the one hand, although the nature of spatial-agnostic along with spatial-compact makes sense in enhancing the efficiency and interpreting the translation equivalence, it deprives convolution kernels of the ability to adapt to diverse visual patterns with respect to different spatial positions. Besides, locality constrains the receptive field of convolution, posing challenges for capturing long-range spatial interactions in a single shot. On the other hand, as is known to us all, inter-channel redundancy inside convolution filters stands out in many successful deep neural networks~\cite{BMVC.28.88}, casting the large flexibility of convolution kernels with respect to different channels into doubt.

To conquer the aforementioned limitations, we present the operation coined as \emph{involution} that has symmetrically \emph{inverse inherent} characteristics compared to convolution, namely, spatial-specific and channel-agnostic. Concretely speaking, involution kernels are distinct in the spatial extent but shared across channels. Being subject to its spatial-specific peculiarity, if involution kernels are parameterized as fixed-sized matrices like convolution kernels and updated using the back-propagation algorithm, the learned involution kernels would be impeded from transferring between input images with variable resolutions. To the end of handling variable feature resolutions, an involution kernel belonging to a specific spatial location is possible to be generated solely conditioned on the incoming feature vector at the corresponding location itself, as an intuitive yet effective instantiation. Besides, we alleviate the redundancy of kernels by sharing the involution kernel along the channel dimension. Taken the above two factors together, the computational complexity of an involution operation scales up linearly with the number of feature channels, based on which an extensive coverage in the spatial dimension is allowed for the dynamically parameterized involution kernels. By virtue of an inverted designing scheme, our proposed involution has two-fold privileges over convolution: ($\romannumeral1$) involution could summarize the context in a wider spatial arrangement, thus overcome the difficulty of modeling long-range interactions well; ($\romannumeral2$) involution could adaptively allocate the weights over different positions, so as to prioritize the most informative visual elements in the spatial domain.

Analogously, recent approaches have spoken for going beyond convolution with the preference of self-attention for the purpose of capturing long-range dependencies~\cite{NIPS2019_8302,Zhao_2020_CVPR}. Among these works, pure self-attention could be utilized to construct stand-alone models with promising performance. Intriguingly, we reveal that self-attention particularizes our generally defined involution through a sophisticated formulation concerning kernel construction. By comparison, the involution kernel adopted in this work is generated conditioned on a single pixel, rather than its relationship with the neighboring pixels. To take one step further, we prove in our experiments that even with our embarrassingly simple version, involution could achieve competitive accuracy-cost trade-offs to self-attention. Being fully aware that the affinity matrix acquired by comparing query with each key in self-attention is also an instantiation of the involution kernel, we question the necessity of composing query and key features to produce such a kernel, since our simplified involution kernel could also attain decent performance while avoiding the superfluous attendance of key content, let alone the dedicated positional encoding in self-attention.

The presented involution operation readily facilitates visual recognition by embedding extendable and switchable spatial modeling into the representation learning paradigm, in a fairly lightweight manner. Built upon this redesigned visual primitive, we establish a backbone architecture family, dubbed as RedNet, which could achieve superior performance over convolution-based ResNet and self-attention based models for image classification. On the downstream tasks including detection and segmentation, we comprehensively perform a step-by-step study to inspect the effectiveness of involution on different components of detectors and segmentors, such as their backbone and neck. Involution is proven to be helpful for each of the considered components, and the combination of them leads to the greatest efficiency.

Summarily, our primary contributions are as follows:

\begin{enumerate}[itemsep=-2pt,topsep=0pt]
	\item We rethink the inherent properties of convolution, associated with the spatial and channel scope. This motivates our advocate of other potential operators embodied with discrimination capability and expressiveness for visual recognition as an alternative, breaking through existing inductive biases of convolution.
	\item We bridge the emerging philosophy of incorporating self-attention into the learning procedure of visual representation. In this context, the desiderata of composing pixel pairs for relation modeling is challenged. Furthermore, we unify the view of self-attention and convolution through the lens of our involution.
	\item The involution-powered architectures work universally well across a wide array of vision tasks, including image classification, object detection, instance and semantic segmentation, offering significantly better performance than the convolution-based counterparts.
\end{enumerate}


\section{Sketch of Convolution}

We initiate from introducing the standard convolution operation to make the definition of our proposed involution self-contained. Let $\varX \in \real^{H \times W \times C_i}$ denote the input feature map, where $H$, $W$ represent its height, width and $C_i$ enumerates the input channels. Inside the cube of a feature tensor $\varX$, each feature vector $\varX_{i,j} \in \real^{C_i}$ located in a cell of the image lattice can be considered as a \emph{pixel} representing certain high-level semantic patterns, with a little abuse of notation.

A cohort of $C_o$ \textbf{convolution filters} with the fixed kernel size of $K \times K$ is denoted as $\varF \in \real^{C_o \times C_i \times K \times K}$, where each filter $\varF_k \in \real^{C_i \times K \times K}, k=1,2,\cdots,C_o$, contains $C_i$ \textbf{convolution kernels} $\varF_{k,c} \in \real^{K \times K}, c=1,2,\cdots,C_i$ and executes Multiply-Add operations on the input feature map in a sliding-window manner to yield the output feature map $\varY \in \real^{H \times W \times C_o}$, defined as
\begin{equation}
	\varY_{i,j,k} = \sum_{c=1}^{C_i} \sum_{(u,v) \in \boldsymbol{\Delta}_K} \varF_{k,c,u+\lfloor K/2 \rfloor,v+\lfloor K/2 \rfloor} \varX_{i+u,j+v,c},
\end{equation}
where $\boldsymbol{\Delta}_K \in \mathbb{Z}^2$ refers to the set of offsets in the neighborhood  considering convolution conducted on the center pixel, written as ($\times$ indicates Cartesian product here)
\begin{equation}
	\boldsymbol{\Delta}_K = \left[-\left\lfloor K/2 \right\rfloor, \cdots, \left\lfloor K/2 \right\rfloor\right] \times \left[-\left\lfloor K/2 \right\rfloor, \cdots, \left\lfloor K/2 \right\rfloor\right].
\end{equation}
Moreover, depth-wise convolution~\cite{Chollet_2017_CVPR} pushes the formulation of group convolution~\cite{NIPS2012_4824,Xie_2017_CVPR} to the extreme, where each filter (virtually degenerated into a single kernel) $\varG_k \in \real^{K \times K},k=1,2,\cdots,C_o$, strictly performs convolution on an individual feature channel indexed by $k$, so the first dimension is eliminated from $\varF_k$ to form $\varG_k$, under the assumption that the number of output channels equals the input ones. As it stands, the convolution operation becomes
\begin{equation}
	\varY_{i,j,k} = \sum_{(u,v) \in \boldsymbol{\Delta}_K} \varG_{k,u+\lfloor K/2 \rfloor,v+\lfloor K/2 \rfloor} \varX_{i+u,j+v,k}.
\end{equation}
Note that the kernel $\varG_k$ is specific to the $k$\textsuperscript{th} feature slice $\varX_{\cdot,\cdot,k}$ from the view of channel and shared among all the spatial locations within this slice.

\section{Design of Involution}

Compared to either standard or depth-wise convolution described above, \textbf{involution kernels} $\varH\in \real^{H \times W \times K \times K \times G}$ are devised to embrace transforms with \textit{inverse} characteristics in the spatial and channel domain, hence its name. Specifically, \textbf{an involution kernel} $\varH_{i,j,\cdot,\cdot,g} \in \real^{K \times K}, g=1,2,\cdots,G$, is specially tailored for the pixel $\varX_{i,j} \in \real^C$  (the subscript of $C$ is omitted for notation brevity) located at the corresponding coordinate $(i, j)$, but shared over the channels. $G$ counts the number of groups where each group shares the same involution kernel. The output feature map of involution is derived by performing Multiply-Add operations on the input  with such involution kernels, defined as
\begin{equation}
	\label{eqn:madd}
	\varY_{i,j,k} = \sum_{(u, v) \in \Delta_K} \boldsymbol{\mathcal{H}}_{i,j,u+\lfloor K/2 \rfloor,v+\lfloor K/2 \rfloor,\lceil kG/C \rceil} \varX_{i+u,j+v,k}.
\end{equation}

\begin{algorithm}[t]
\caption{\small{Pseudo code of involution in a PyTorch-like style.}}
\label{alg:code}
\definecolor{codeblue}{rgb}{0.25,0.5,0.5}
\lstset{
	backgroundcolor=\color{white},
	basicstyle=\fontsize{7.2pt}{7.2pt}\ttfamily\selectfont,
	columns=fullflexible,
	breaklines=true,
	captionpos=b,
	commentstyle=\fontsize{7.2pt}{7.2pt}\color{codeblue},
	keywordstyle=\fontsize{7.2pt}{7.2pt},
}
\vskip -0.075in
\begin{lstlisting}[language=python]
# B: batch size, H: height, W: width
# C: channel number, G: group number
# K: kernel size, s: stride, r: reduction ratio

################### initialization ###################
o = nn.AvgPool2d(s, s) if s > 1 else nn.Identity()
reduce = nn.Conv2d(C, C//r, 1)
span = nn.Conv2d(C//r, K*K*G, 1)
unfold = nn.Unfold(K, dilation, padding, s)
#################### forward pass ####################
x_unfolded = unfold(x) # B,CxKxK,HxW
x_unfolded = x_unfolded.view(B, G, C//G, K*K, H, W)
# kernel generation, Eqn.(6)
kernel = span(reduce(o(x))) # B,KxKxG,H,W
kernel = kernel.view(B, G, K*K, H, W).unsqueeze(2)
# Multiply-Add operation, Eqn.(4)
out = mul(kernel, x_unfolded).sum(dim=3) # B,G,C/G,H,W
out = out.view(B, C, H, W)
return out
\end{lstlisting}
\vskip -0.075in
\end{algorithm}

Different from convolution kernels, the shape of involution kernels $\varH$ depends on that of the input feature map $\varX$. A natural thought is to generate the involution kernels conditioned on (part of) the original input tensor, so that the output kernels would be comfortably aligned to the input. We symbolize the kernel generation function as $\localFunc$ and abstract the functional mapping at each location $(i,j)$ as
\begin{equation}
	\label{eqn:kernel-gen}
	\varH_{i,j} = \localFunc(\varX_{\boldsymbol{\Psi}_{i,j}}),
\end{equation}
where $\boldsymbol{\Psi}_{i,j}$ indexes the set of pixels $\varH_{i,j}$ is conditioned on.

\begin{figure}[t]
	\vskip -0.325in
	\begin{center}
		\includegraphics[width=\linewidth,trim=5 5 5 5,clip]{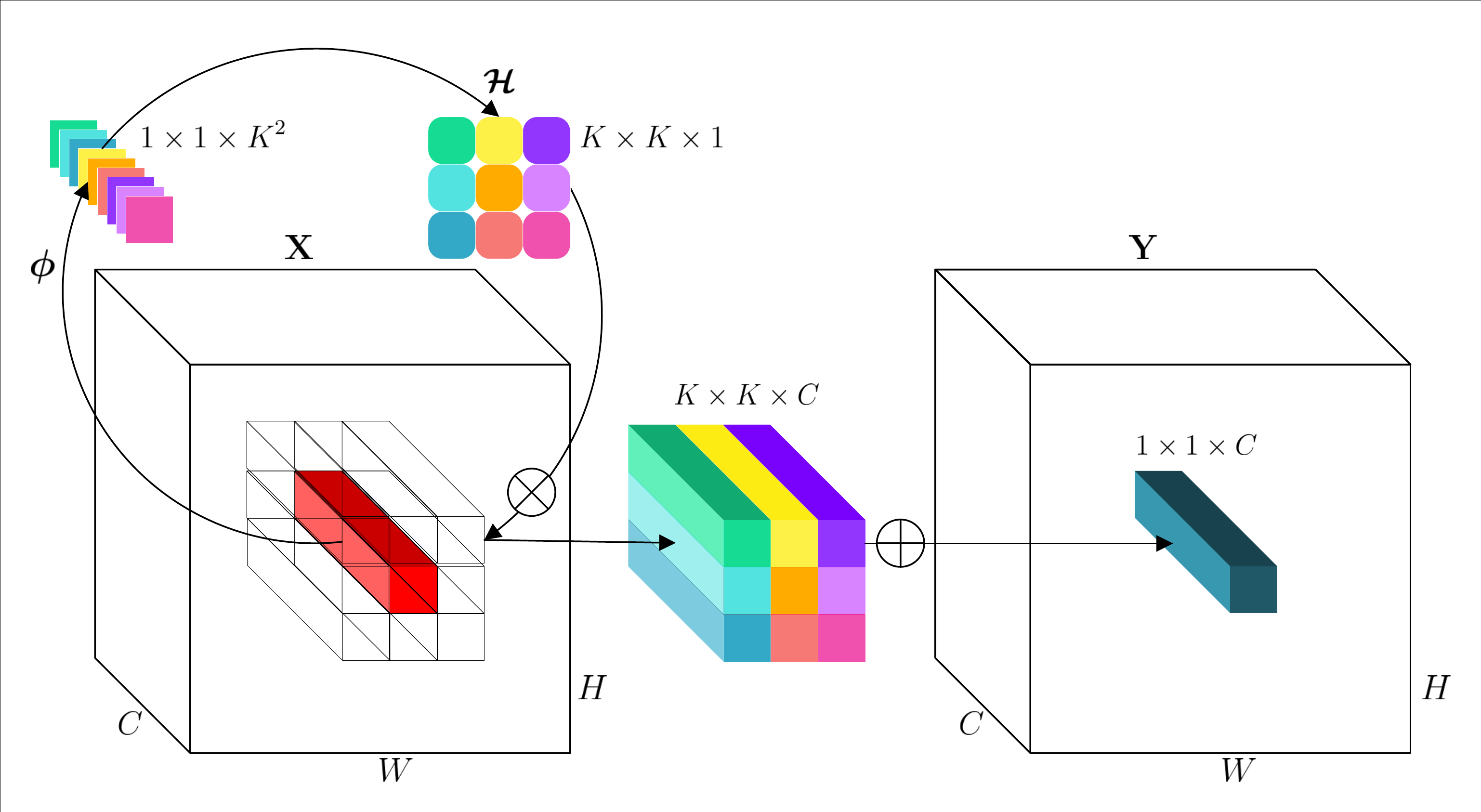}
	\end{center}
	\vskip -0.275in
	\caption{Schematic illustration of our proposed involution. The involution kernel $\varH_{i,j} \in \real^{K \times K \times 1}$ ($G=1$ in this example for ease of demonstration) is yielded from the function $\localFunc$ conditioned on a single pixel at $(i,j)$, followed by a channel-to-space rearrangement. The Multiply-Add operation of involution is decomposed into two steps, with $\bigotimes$ indicating multiplication broadcast across $C$ channels and $\bigoplus$ indicating summation aggregated within the  $K \times K$ spatial neighborhood. Best viewed in color.}
	\label{fig:involution}
	\vskip -0.2in
\end{figure}

\vspace{-1.0em}
\paragraph{Implementation Details} Respectful of the conciseness of convolution, we make involution conceptually as simple as possible. Note that our target is to firstly provide a design space for the kernel generation function $\localFunc$ and then fast prototype some effective designing instances for practical usage. In this work, we choose to span each involution kernel $\varH_{i,j}$ from a single pixel $\varX_{i,j}$ for incarnation. More exquisite designs under exploration may have the potential of further pushing the performance boundary, but are left as future work. Besides, we are conscious that self-attention falls into this design space while being a more complicated materialization than our default choice, which is to be discussed in more detail in Section~\ref{sec:attention}. Formally, we have the kernel generation function $\localFunc \colon \real^{C} \mapsto \real^{K \times K \times G}$ with $\boldsymbol{\Psi}_{i,j} = \{(i,j)\}$ taking the following form:
\begin{equation}
	\label{eqn:pixel-wise-gen}
	\varH_{i,j} = \localFunc(\varX_{i,j}) = \mathbf{W}_1 \sigma(\mathbf{W}_0 \varX_{i,j}).
\end{equation}
In this formula, $\mathbf{W}_0 \in \real^{\frac{C}{r} \times C}$ and $\mathbf{W}_1 \in \real^{(K \times K \times G) \times \frac{C}{r}}$ represent two linear transformations that collectively constitute a bottleneck structure, where the intermediate channel dimension is under the control of a reduction ratio $r$ for efficient processing, and $\sigma$ implies Batch Normalization and non-linear activation functions that interleave two linear projections. We refer to Eqn.~\ref{eqn:madd} with the materialized kernel generation function of Eqn.~\ref{eqn:pixel-wise-gen} as involution hereinafter. The pseudo code shown in Alg.~\ref{alg:code} delineates the computation flow of involution, which is visualized in Figure~\ref{fig:involution}.

For building the entire network with involution, we mirror the design of ResNet~\cite{He_2016_CVPR} by stacking residual blocks, since the elegant architecture of ResNet makes it apt for incubating new ideas and making comparisons. We replace involution for $3 \times 3$ convolution at all bottleneck positions in the stem (using $3 \times 3$ or $7 \times 7$ involution for classification or dense prediction) and trunk (using $7 \times 7$  involution for all tasks) of ResNet, but retain all the $1 \times 1$ convolution for channel projection and fusion. These delicately redesigned entities unite to shape a new species of highly efficient backbone networks, termed as RedNet.

Once spatial and channel information interweaves, heavy redundancy tends to occur inside the neural networks. However, the information interactions are tactfully decoupled in our RedNet towards a favorable accuracy-efficiency trade-off, as empirically evidenced in Figure~\ref{fig:efficiency}. To be specific, the information encoded in the channel dimension of one pixel is implicitly scattered to its spatial vicinity in the kernel generation step, after which the information in an enriched receptive field is gathered thanks to the vast and dynamic involution kernels. Indispensably, linear transformations (realized by $1 \times 1$ convolutions) are interspersed for channel information exchange. In a word, channel-spatial, spatial-alone, and channel-alone interactions alternately and independently act on the stream of information propagation, collaboratively facilitating the miniaturization of network architectures while ensuring the representation capability.

\section{In Context of Prior Literature}

This section relates to several important aspects revolving around neural architecture in prior literature. We clarify their similarities and differences compared to our method.

\subsection{Convolution and Variants} 
As the \textit{de facto} standard operator of modern vision systems, convolution~\cite{726791} possesses two principal characteristics, spatial-agnostic and channel-specific. Convolution kernels are location-independent in the spatial extent for translation equivalence but privatized at different channels for information discrimination. Along another research line, depth-wise convolution demonstrates wide applicability in efficient neural network architecture design~\cite{Chollet_2017_CVPR,Sandler_2018_CVPR,Ma_2018_ECCV,pmlr-v97-tan19a}.  The depth-wise convolution is a pioneering attempt towards factorizing the spatial and channel entanglement of standard convolution, which is symmetric to our proposed involution operation in that depth-wise convolution contains a set of kernels specific to each channel and spatially-shared while our invented involution kernels are shared over channels and dedicated to each planar location in the image lattice.

Until most recently, dynamic convolutions emerge as powerful variants of the stationary ones. These approaches either straightforwardly generate the entire convolution filters~\cite{Ha2017Hypernetworks,NIPS2016_6578,NIPS2019_8412}, or parameterize the sampling grid associated with each convolution kernel~\cite{Dai_2017_ICCV,Jeon_2017_CVPR,Zhu_2019_CVPR}. Regarding the former category~\cite{Ha2017Hypernetworks,NIPS2016_6578,NIPS2019_8412}, unlike us, their dynamically generated convolution filters still conform to the two properties of standard convolution, thus incurring significant memory or computation consumption for filter generation. Regarding the latter category~\cite{Dai_2017_ICCV,Jeon_2017_CVPR,Zhu_2019_CVPR}, only certain attributes, \eg, the footprint of convolution kernels, are determined in an adaptive fashion.

Actually, early in the field of face recognition, DeepFace~\cite{Taigman_2014_CVPR} and DeepID~\cite{Sun_2014_CVPR} have explored locally connected layers without weight sharing in the spatial domain, enlightened by apparently different regional distributions of statistics in the face imagery. Nevertheless, such excessive relaxation of convolution parameters can be problematic in knowledge transfer from one position to others. Resembling dynamic convolutions, our involution tackles this dilemma through sharing \emph{meta-weights} of the kernel generation function across different positions, though not directly the \emph{weights} of kernel instances. There also exist previous works that adopt pixel-wise dynamic kernels for feature aggregation, but they mainly capitalize on the context information for feature up-sampling~\cite{Su_2019_CVPR,Wang_2019_ICCV} and still rely on convolution for basic feature extraction. The most relevant work towards substituting convolution rather than up-sampling might be~\cite{Esquivel_2019_ICCV}, but the pixel-wise generated filters still inherit one original property of convolution, to perform feature aggregation in a distinct manner over each channel.

\subsection{Attention Mechanism}
\label{sec:attention}
The attention mechanism originates from the field of machine translation~\cite{NIPS2017_7181} and exhibits blossoming development in the arena of natural language processing~\cite{dai-etal-2019-transformer,NIPS2019_8812}. Its success has also been translated to a plethora of vision tasks, including image recognition~\cite{Bello_2019_ICCV,Hu_2019_ICCV,NIPS2019_8302,Zhao_2020_CVPR}, image generation~\cite{pmlr-v80-parmar18a,pmlr-v97-zhang19d}, video understanding~\cite{Sun_2019_ICCV,Wang_2018_CVPR}, object detection~\cite{Carion_2020_ECCV,Hu_2018_CVPR,Zhu_2019_ICCV}, and semantic segmentation~\cite{Fu_2019_CVPR,Huang_2019_ICCV,Wang_2020_ECCV}. Some works sparingly insert self-attention as plugin modules into the backbone neural network~\cite{NIPS2018_7318,NIPS2018_7886} or attach them on the top of the backbone to extract high-level semantic relationships~\cite{Carion_2020_ECCV,Sun_2019_ICCV}, retaining the substratum of convolutional features. More aggressively, other works adopt the off-the-shell self-attention layer as the fundamental backbone component for vision~\cite{Bello_2019_ICCV,Hu_2019_ICCV,NIPS2019_8302,Wang_2020_ECCV,Zhao_2020_CVPR}. Still, limited emphasis has been laid on delving deep into the learning dynamics of this functional form compared to convolution~\cite{Cordonnier2020On}.

Our proposed involution in Eqn.~\ref{eqn:madd} is reminiscent of self-attention and essentially could become a generalized version of it. The self-attention pools \emph{values} $\varV$ depending on the affinities obtained by computing correspondences between the \emph{query} and \emph{key} content, $\varQ$ and $\varK$, formulized as
\begin{equation}
	\label{eqn:self-attn}
	\varY_{i,j,k} = \sum_{(p, q) \in \boldsymbol{\Omega}} (\varQ \varK^\top)_{i,j,p,q,\lceil kH/C \rceil} \varV_{p,q,k},
\end{equation}
where $\varQ$, $\varK$ and $\varV$ are linearly transformed from the input $\varX$, and $H$ is the number of heads in multi-head self-attention~\cite{NIPS2017_7181}. The similarity lies in that both operators collect pixels in the neighborhood $\boldsymbol{\Delta}$ or a less bounded scope $\boldsymbol{\Omega}$ through a weighted sum. On the one hand, the computing regime of involution can be considered as an attentive aggregation over the spatial domain. On the other hand, the attention map, or say affinity matrix $\varQ\varK^\top$ in the self-attention, can be viewed as a kind of involution kernel $\varH$.

However, with the particulars of kernel generation comes the differences between self-attention and our materialized involution form with Eqn.~\ref{eqn:pixel-wise-gen}. Regrading previous endeavor on replacing convolution with local self-attention~\cite{Hu_2019_ICCV,NIPS2019_8302,Zhao_2020_CVPR} to establish backbone models, they have to derive the affinity matrix (equivalent to involution kernel in our context) based on the relationship between the \emph{query} and \emph{key content}, optionally with hand-crafted \emph{relative positional encoding} for permutation-variance. From this point of view, for self-attention, the input to the kernel generation function in Eqn.~\ref{eqn:kernel-gen} would become a set of pixels indexed by $\boldsymbol{\Psi}_{i,j} = (i,j) + \boldsymbol{\Delta}_K$\footnote{$+$ indicates adding a variable vector to each element in a set here.}, including both the pixel of interest and its surrounding ones. Subsequently, the function could compose all these attended pixels, in an either ordered~\cite{Zhao_2020_CVPR} or unordered~\cite{Hu_2019_ICCV,NIPS2019_8302,Zhao_2020_CVPR} manner, and exploit complex relationships between them. In stark contrast to above, we constitute the involution kernel via operating solely on the original input pixel itself with $\boldsymbol{\Psi}_{i,j} = \{(i,j)\}$, as expressed by Eqn.~\ref{eqn:pixel-wise-gen}. From the perspective of self-attention, our involution kernels only explicitly rely on the \emph{query content}, while the \emph{relative positional information} is implicitly encoded in the organized output form of our kernel generation function. We sacrifice the pixel-paired relationship modeling, but the final performance of our RedNet is on par with those heavily relation-based models. Therefore, we may reach a conclusion that it is the macro design principles of involution instead of its micro setup nuances that are instrumental in the representation learning for visual understanding, corroborated by our empirical results in the experimental part. Another strong evidence supporting our hypothesis is that only using position encoding (by replacing $\varQ\varK^\top$ in Eqn.~\ref{eqn:self-attn} with $\varQ\varR^\top$, where $\varR$ is the position embedding matrix) retains descent performance of self-attention based models~\cite{NIPS2019_8302,bello2021lambdanetworks}. Previously, the above observation is interpreted as the crucial role of position encoding in self-attention, but now a reinterpretation of the root cause behind might be $\varQ\varR^\top$ is still a form of dynamically parameterized involution kernel.

More importantly, precedent self-attention based works seldom show their versatility in multifarious vision tasks, but our involution paves a viable pathway for a great variety of tasks, as we shall find soon in Section~\ref{sec:main-exp}.


\section{Experiments}

\subsection{Main Results}
\label{sec:main-exp}
We conduct comprehensive experiments from conceptual prediction to (semi-)dense prediction. All the network models are implemented with the PyTorch library~\cite{NIPS2019_9015}.

\subsubsection{Image Classification}

We perform the backbone training from scratch on the ImageNet~\cite{imagenet_cvpr09} training set that is one of the most challenging benchmarks for object recognition up to date. For a fair comparison, we adhere to the training protocol of Stand-Alone Self-Attention~\cite{NIPS2019_8302} and Axial Attention~\cite{Wang_2020_ECCV}, except that we \emph{do not} use exponential moving average (EMA) over the trainable parameters during training. Following the identical recipe, we re-implement both pairwise and patchwise SAN~\cite{Zhao_2020_CVPR} with their open-source code\footnote{\url{https://github.com/hszhao/SAN}} as a stronger baseline, and show our reproduced results in the corresponding tables and figures respectively.
The detailed training setup is provided in the Appendix. We apply the Inception-style preprocessing for data augmentation~\cite{Szegedy_2015_CVPR}, \ie, random resized cropping and horizontal flipping. For evaluation, we use the single-crop testing method on the validation set following the common practice.

\begin{table}[t]
	\centering
	\tablestyle{5pt}{1.0}
	\resizebox{.9\linewidth}{!}{
		\begin{tabular}{c|c|c|c}
			\fontsize{7pt}{1em}\selectfont Architecture 
			& \fontsize{7pt}{1em}\selectfont \#Params (M)
			& \fontsize{7pt}{1em}\selectfont FLOPs (G)
			& \fontsize{7pt}{1em}\selectfont Top-1 Acc. (\%) \\
			\shline
			ResNet-26~\cite{He_2016_CVPR} & 13.7 & 2.4 & 73.6 \\
			LR-Net-26~\cite{Hu_2019_ICCV} & 14.7 & 2.6 & 75.7 \\
			Stand-Alone ResNet-26~\cite{NIPS2019_8302} & 10.3 & 2.4 & 74.8 \\
			SAN10$^\dagger$~\cite{Zhao_2020_CVPR} & 10.5 & 2.2 & 75.5 \\
			\rowcolor{cyan!20}
			RedNet-26 & \textbf{9.2} & \textbf{1.7} & \textbf{75.9} \\
			\hline\hline
			ResNet-38~\cite{He_2016_CVPR} & 19.6 & 3.2 & 76.0 \\
			Stand-Alone ResNet-38~\cite{NIPS2019_8302} & 14.1 & 3.0 & 76.9 \\
			SAN15$^\dagger$~\cite{Zhao_2020_CVPR} & 14.1 & 3.0 & 77.1 \\
			\rowcolor{cyan!20}
			RedNet-38 & \textbf{12.4} & \textbf{2.2} & \textbf{77.6} \\
			\hline\hline
			ResNet-50~\cite{He_2016_CVPR} & 25.6 & 4.1 & 76.8 \\
			LR-Net-50~\cite{Hu_2019_ICCV} & 23.3 & 4.3 & 77.3 \\
			AA-ResNet-50~\cite{Bello_2019_ICCV} & 25.8 & 4.2 & 77.7 \\
			Stand-Alone ResNet-50~\cite{NIPS2019_8302} & 18.0 & 3.6 & 77.6 \\
			SAN19$^\dagger$~\cite{Zhao_2020_CVPR} & 17.6 & 3.8 & 77.4 \\
			Axial ResNet-S$^\ddagger$~\cite{Wang_2020_ECCV} & \textbf{12.5} & 3.3 & 78.1 \\
			\rowcolor{cyan!20}
			RedNet-50 & 15.5 & \textbf{2.7} & \textbf{78.4} \\
			\hline\hline
			ResNet-101~\cite{He_2016_CVPR} & 44.6 & 7.9 & 78.5 \\
			LR-Net-101~\cite{Hu_2019_ICCV} & 42.0 & 8.0 & 78.5 \\
			AA-ResNet-101~\cite{Bello_2019_ICCV} & 45.4 & 8.1 & 78.7 \\
			\rowcolor{cyan!20}
			RedNet-101 & \textbf{25.6} & \textbf{4.7} & \textbf{79.1} \\
			\hline\hline
			ResNet-152~\cite{He_2016_CVPR} & 60.2 & 11.6 & \textbf{79.3} \\
			AA-ResNet-152~\cite{Bello_2019_ICCV} & 61.6 & 11.9 & 79.1 \\
			Axial ResNet-M$^\ddagger$~\cite{Wang_2020_ECCV} & \textbf{26.5} & \textbf{6.8} & 79.2 \\
			Axial ResNet-L$^\ddagger$~\cite{Wang_2020_ECCV} & 45.8 & 11.6 & \textbf{79.3} \\
			\rowcolor{cyan!20}
			RedNet-152 & 34.0 & \textbf{6.8} & \textbf{79.3} \
		\end{tabular}
	}
	\vskip -0.1in
	\caption{The architecture profiles on ImageNet val set. Single-crop testing with $224 \times 224$ crop size is adopted. We compare with improved re-implementations if available and extract the other results from their original publications.
		\newline
		{\fontsize{7.2pt}{0pt}\selectfont{
				$^\dagger$ \emph{The improved re-implementation results of pairwise SAN models are listed here.
		}}}
		\newline
		{\fontsize{7.2pt}{0pt}\selectfont{
				$^\ddagger$ \emph{Axial ResNet modifies the architecture setup of ResNet by changing the reduction ratio in each bottleneck block from 4 to 2.
		}}}
	} 
	\label{tab:cls}
	\vskip -0.15in
\end{table}

\begin{table}[t]
	\centering
	\tablestyle{5pt}{1.0}
	\resizebox{.9\linewidth}{!}{
	\begin{tabular}{c|c|c|c}
		\fontsize{7pt}{1em}\selectfont Architecture 
		& \fontsize{7pt}{1em}\selectfont GPU time (ms)
		& \fontsize{7pt}{1em}\selectfont CPU time (ms)
		& \fontsize{7pt}{1em}\selectfont Top-1 Acc. (\%) \\
		\shline
		ResNet-50~\cite{He_2016_CVPR} & 11.4 & 895.4 & 76.8 \\
		ResNet-101~\cite{He_2016_CVPR} & 18.9 & 967.4 & 78.5 \\
		SAN19~\cite{Zhao_2020_CVPR} & 33.2 & N/A & 77.4 \\
		Axial ResNet-S~\cite{Wang_2020_ECCV} & 35.9 & 377.0 & 78.1 \\
		\hline
		RedNet-38 & 11.4 & 156.3 & 77.6 \\
		RedNet-50 & 14.3 & 211.2 & 78.4 \\
	\end{tabular}
	}
	\vskip -0.1in
	\caption{Runtime analysis for representative networks. The speed benchmark is on a single NVIDIA TITAN Xp GPU and Intel\textsuperscript{\textregistered} Xeon\textsuperscript{\textregistered} CPU E5-2660 v4@2.00GHz.
	} 
	\label{tab:runtime}
	\vskip -0.2in
\end{table}

\begin{figure*}[t]
	\vskip -0.1in
	\begin{center}
		\subfloat[The accuracy-complexity envelope on ImageNet.]{
			\begin{minipage}{.45\linewidth}
				\centering
				\vskip -0.1in
				\includegraphics[width=\linewidth]{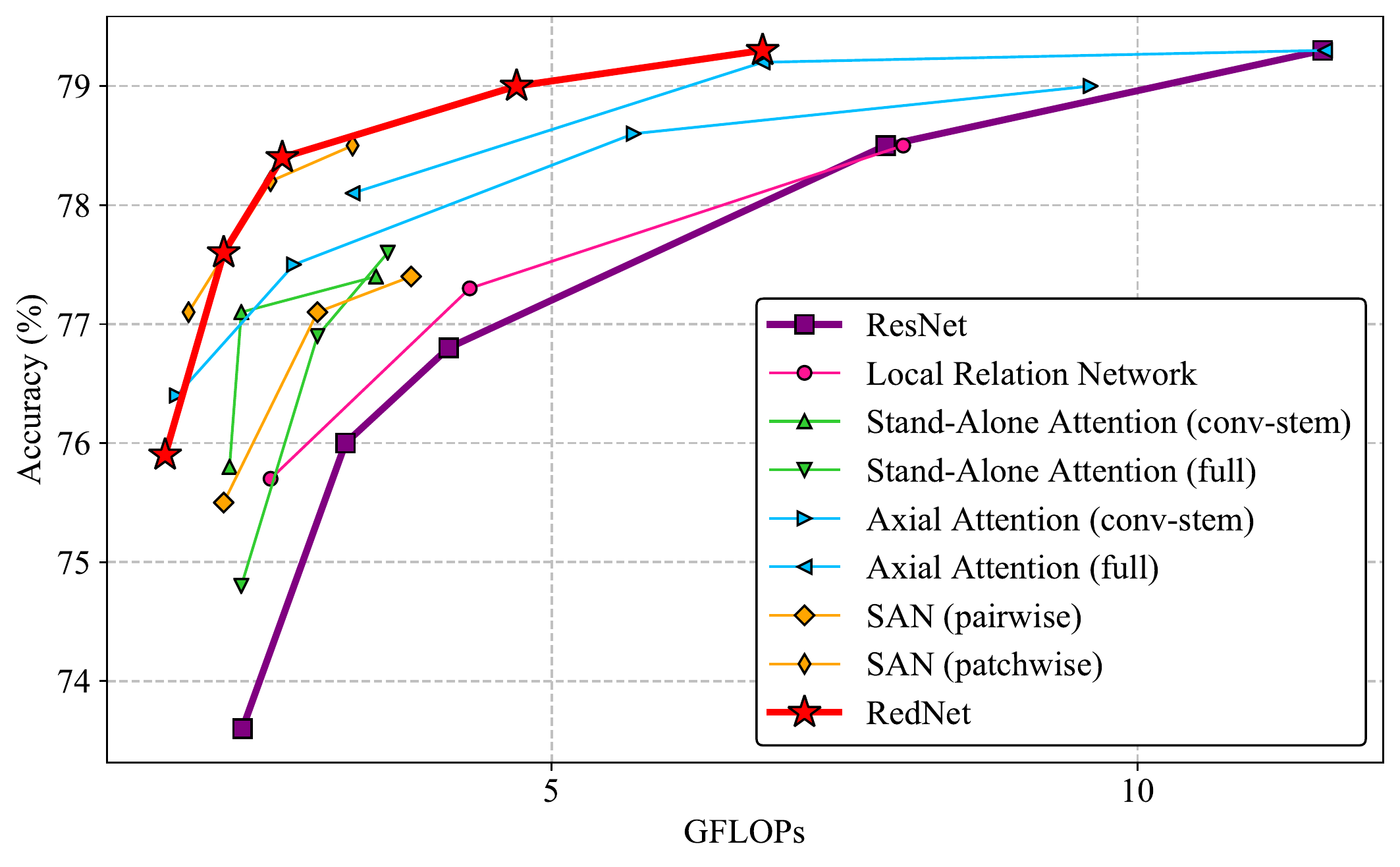}
				\vskip -0.01in
				\label{fig:complexity}
		\end{minipage}}
		\quad\quad\quad
		\subfloat[The accuracy-parameter envelope on ImageNet.]{
			\begin{minipage}{.45\linewidth}
				\centering
				\vskip -0.1in
				\includegraphics[width=\linewidth]{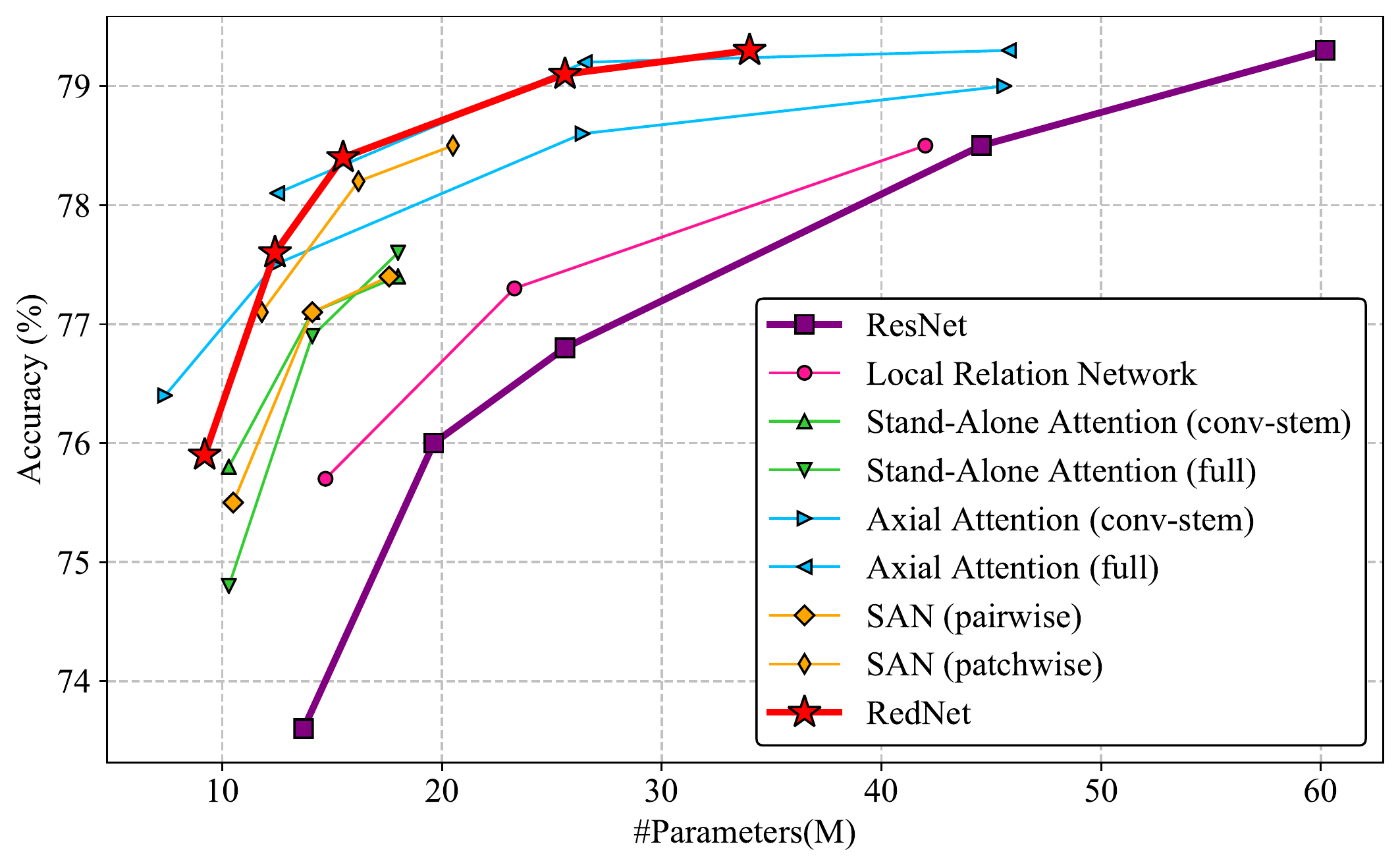}
				\vskip -0.01in
				\label{fig:parameter}
		\end{minipage}}
	\end{center}
	\vskip -0.2in
	\caption{The accuracy-efficiency envelopes on ImageNet val set. This figure visualizes Table~\ref{tab:cls}. In general, RedNet achieves the optimal trade-off in comparison with all the other convolution and self-attention based architectures.}
	\label{fig:efficiency}
	\vskip -0.2in
\end{figure*}

In the same spirit of ResNet, we scale the network depth to establish our RedNet family. The comparison to convolution and self-attention based vision models are summarized in Table~\ref{tab:cls}. Almost within each group of the table, RedNet achieves the highest recognition accuracy whilst with the most parsimonious parameter storage and computational budget. RedNet could substantially outperform ResNet across all depths. For example, RedNet-50 achieves a margin of 1.6\% higher accuracy over ResNet-50, using 39.5\% fewer parameters and 34.1\% lower computational consumption. Moreover, RedNet-50 is on par with ResNet-101 regrading to the top-1 recognition accuracy, while saving 65.2\% and 65.8\% storage and computation resources respectively. For an intuitive demonstration, the corresponding accuracy-complexity envelope is illustrated in Figure~\ref{fig:complexity}, where our RedNet shows the top-performing Pareto frontier, in abreast with other state-of-the-art self-attention models, while being free from more complex relation modeling. Likewise, we could observe a similar trend in the accuracy-parameter envelope shown in Figure~\ref{fig:parameter}. It is noteworthy that RedNet strikes a better balance between parameters and complexities, compared to the top competitors like SAN and Axial ResNet, as they are enveloped by the curve of RedNet series either in Figure~\ref{fig:complexity} or~\ref{fig:parameter}.

To reflect the practical runtime, we measure the inference time of different architectures with the comparable performance for a single image with the shape of $224 \times 224$. We report the running time on GPU/CPU in Table~\ref{tab:runtime}, where RedNet demonstrates its merits in terms of wall-clock time under the same level of accuracy. A customized CUDA kernel implementation with optimized memory scheduling for involution is highly anticipated for further acceleration on GPU. Depending on the extent to which optimizing hardware accelerators is contributed to this new involution operator, on-device speedup might approach the theoretical speedup compared to convolution in the future.

\begin{table*}[t]
\centering
\tablestyle{4pt}{1.0}
\resizebox{.9\linewidth}{!}{
	\begin{tabular}{c|c|c|c|c|l|l|l|l|l|l}
		\fontsize{7pt}{1em}\selectfont Detector & \fontsize{7pt}{1em}\selectfont Backbone & \fontsize{7pt}{1em}\selectfont Neck
		& \fontsize{7pt}{1em}\selectfont \#Params (M)
		& \fontsize{7pt}{1em}\selectfont FLOPs (G)
		& \fontsize{7pt}{1em}\selectfont \quad\apbbox{~}
		& \fontsize{7pt}{1em}\selectfont \quad\apbbox{50}
		& \fontsize{7pt}{1em}\selectfont \quad\apbbox{75}
		& \fontsize{7pt}{1em}\selectfont \quad\apbbox{S}
		& \fontsize{7pt}{1em}\selectfont \quad\apbbox{M}
		& \fontsize{7pt}{1em}\selectfont \quad\apbbox{L} \\
		\shline
		\multirow{3}{*}{RetinaNet~\cite{Lin_2017_ICCV}} & ResNet-50 & convolution & 37.7 & 239.3 & 36.6 & 55.8 & 39.2 & 20.9 & 40.6 & 47.5 \\
		& RedNet-50 & convolution & 27.8 & 210.1 & 38.3\cgap{+}{1.7} & 58.2\cgaphl{+}{2.4} & 40.5\cgap{+}{1.3} & 21.1\cgap{+}{0.2} & 41.8\cgap{+}{1.2} & 50.9\cgaphl{+}{3.4} \\
		& RedNet-50 & involution & 26.3 & 199.9 & 38.2\cgap{+}{1.6} & 58.2\cgaphl{+}{2.4} & 40.4\cgap{+}{1.2} & 21.8\cgap{+}{0.9} & 41.6\cgap{+}{1.0} & 50.7\cgaphl{+}{3.2} \\
	\end{tabular}
}
\vskip 0.05in
\resizebox{\linewidth}{!}{
	\begin{tabular}{c|c|c|c|c|c|l|l|l|l|l|l}
		\fontsize{7pt}{1em}\selectfont Detector & \fontsize{7pt}{1em}\selectfont Backbone & \fontsize{7pt}{1em}\selectfont Neck & \fontsize{7pt}{1em}\selectfont Head
		& \fontsize{7pt}{1em}\selectfont \#Params (M)
		& \fontsize{7pt}{1em}\selectfont FLOPs (G)
		& \fontsize{7pt}{1em}\selectfont \quad\apbbox{~}
		& \fontsize{7pt}{1em}\selectfont \quad\apbbox{50}
		& \fontsize{7pt}{1em}\selectfont \quad\apbbox{75}
		& \fontsize{7pt}{1em}\selectfont \quad\apbbox{S}
		& \fontsize{7pt}{1em}\selectfont \quad\apbbox{M}
		& \fontsize{7pt}{1em}\selectfont \quad\apbbox{L} \\
		\shline
		\multirow{4}{*}{Faster R-CNN~\cite{NIPS2015_5638}} & ResNet-50 & convolution & convolution & 41.5 & 207.1 & 37.7 & 58.7 & 40.8 & 21.7 & 41.6 & 48.4 \\
		& RedNet-50 & convolution & convolution & 31.6 & 177.9 & 39.5\cgap{+}{1.8} & 60.9\cgaphl{+}{2.2} & 42.8\cgaphl{+}{2.0} & 23.3\cgap{+}{1.6} & 42.9\cgap{+}{1.3} & 52.2\cgaphl{+}{3.8} \\
		& RedNet-50 & involution & convolution & 29.5 & 135.0 & 40.2\cgaphl{+}{2.5} & 62.1\cgaphl{+}{3.4} & 43.4\cgaphl{+}{2.6} & 24.2\cgaphl{+}{2.5} & 43.3\cgap{+}{1.7} & 52.7\cgaphl{+}{4.3} \\
		& RedNet-50 & involution & involution & 29.0 & 91.5 & 39.2\cgap{+}{1.5} & 61.0\cgaphl{+}{2.3} & 42.4\cgap{+}{1.6} & 23.1\cgap{+}{1.4} & 43.0\cgap{+}{1.4} & 50.7\cgaphl{+}{2.3} \\
	\end{tabular}
}
\vskip 0.05in
\resizebox{\linewidth}{!}{
	\begin{tabular}{c|c|c|c|c|c|l|l|l|l|l|l}
		\fontsize{7pt}{1em}\selectfont Detector & \fontsize{7pt}{1em}\selectfont Backbone & \fontsize{7pt}{1em}\selectfont Neck & \fontsize{7pt}{1em}\selectfont Head
		& \fontsize{7pt}{1em}\selectfont \#Params (M)
		& \fontsize{7pt}{1em}\selectfont FLOPs (G)
		& \fontsize{7pt}{1em}\selectfont ~\quad\ap{~}
		& \fontsize{7pt}{1em}\selectfont ~\quad\ap{50}
		& \fontsize{7pt}{1em}\selectfont ~\quad\ap{75}
		& \fontsize{7pt}{1em}\selectfont ~\quad\ap{S}
		& \fontsize{7pt}{1em}\selectfont ~\quad\ap{M}
		& \fontsize{7pt}{1em}\selectfont ~\quad\ap{L} \\
		\shline
		\multirow{8}{*}{~Mask R-CNN~\cite{He_2017_ICCV}} & \multirow{2}{*}{ResNet-50} & \multirow{2}{*}{convolution} & \multirow{2}{*}{convolution} & \multirow{2}{*}{44.2} & \multirow{2}{*}{253.4} & 38.4 & 59.2 & 41.9 & 21.9 & 42.3 & 49.7 \\
		& & & & & & 35.1 & 56.3 & 37.3 & 18.5 & 38.6 & 46.9 \\
		\cline{2-12}
		& \multirow{2}{*}{RedNet-50} & \multirow{2}{*}{convolution} & \multirow{2}{*}{convolution} & \multirow{2}{*}{34.2} & \multirow{2}{*}{224.2} & 40.2\cgap{+}{1.8} & 61.4\cgaphl{+}{2.2} & 43.7\cgap{+}{1.8} & 24.2\cgaphl{+}{2.3} & 43.4\cgap{+}{1.1} & 52.5\cgaphl{+}{2.8} \\
		& & & & & & 36.1\cgap{+}{1.0} & 58.1\cgap{+}{1.8} & 38.2\cgap{+}{0.9} & 19.9\cgap{+}{1.4} & 39.3\cgap{+}{0.7} & 48.9\cgaphl{+}{2.0} \\
		\cline{2-12}
		& \multirow{2}{*}{RedNet-50} & \multirow{2}{*}{involution} & \multirow{2}{*}{convolution} & \multirow{2}{*}{32.2} & \multirow{2}{*}{181.3} & 40.8\cgaphl{+}{2.4} & 62.3\cgaphl{+}{3.1} & 44.3\cgaphl{+}{2.4} & 24.2\cgaphl{+}{2.3} & 44.0\cgap{+}{1.7} & 53.0\cgaphl{+}{3.3} \\
		& & & & & & 36.4\cgap{+}{1.3} & 59.0\cgaphl{+}{2.7} & 38.5\cgap{+}{1.2} & 19.9\cgap{+}{1.4} & 39.4\cgap{+}{0.8} & 49.1\cgaphl{+}{2.2} \\
		\cline{2-12}
		& \multirow{2}{*}{RedNet-50} & \multirow{2}{*}{involution} & \multirow{2}{*}{involution} & \multirow{2}{*}{29.5} & \multirow{2}{*}{104.6} & 39.6\cgap{+}{1.2} & 60.7\cgap{+}{1.5} & 42.7\cgap{+}{0.8} & 23.5\cgap{+}{1.6} & 43.1\cgap{+}{0.8} & 51.1\cgap{+}{1.4} \\
		& & & & & & 35.1\cgap{+}{0.0} & 57.1\cgap{+}{0.8} & 37.3\cgap{+}{0.0} & 19.2\cgap{+}{0.7} & 38.5\cgap{-}{0.1} & 47.3\cgap{+}{0.4} \\
	\end{tabular}
}
\vskip -0.1in
\caption{Performance comparison on COCO detection and segmentation. The bounding box AP is reported for the object detection track in the upper table. The bounding box and mask AP are simultaneously reported for the instance segmentation track in the lower table, listed in the two separate lines following a single detector. In the parentheses are the gaps to the fully convolution-based counterparts. Highlighted in green are the gaps of at least {\fontsize{8pt}{1em}\selectfont \hl{${+}$\textbf{2.0}}} points, the same in Table~\ref{tab:seg-fpn} and~\ref{tab:seg-upernet}.
} 
\label{tab:det}
\vskip -0.1in
\end{table*}

\vspace{-0.5em}
\subsubsection{Object Detection and Instance Segmentation}

Beyond fundamental image classification, we demonstrate the generalization ability of our proposed involution on downstream vision tasks, such as object detection and instance segmentation. For object detection, we employ the representative one- and two-stage detectors, RetinaNet~\cite{Lin_2017_ICCV} and Faster R-CNN~\cite{NIPS2015_5638}, both equipped with the FPN~\cite{Lin_2017_CVPR} neck. For instance segmentation, we adopt the main-stream detection system, Mask R-CNN~\cite{He_2017_ICCV}, also in companion with FPN. These three detectors with the underlying backbones, ResNet-50 or RedNet-50, are fine-tuned on the Microsoft COCO~\cite{10.1007/978-3-319-10602-1_48} \texttt{train2017} set for transferring the learned representations of images. More training details are reported in the appendix. During quantitative evaluation, we test on the \texttt{val2017} set and report the COCO-style mean Average Precision (mAP) under different IoU thresholds ranging from 0.5 to 0.95 with an increment of 0.05.

Table~\ref{tab:det} compares our models against the baseline of ResNet backbone with the convolution-based neck and head. First, with the RedNet backbone, all the three detectors excel their ResNet-based counterparts with considerable performance gains, \ie, 1.7\%, 1.8\%, and 1.8\% higher in bounding box AP, while being more parameter- and computation-conserving. Second, additionally swapping involution for convolution in the FPN neck brings about healthy margins for Faster/Mask R-CNN, while further reducing their parameters and computational cost to 71\%/73\% and 65\%/72\%. In particular, the margins with respect to bounding box AP are enlarged to 2.5\% and 2.4\% respectively. Third, to build fully involution-based detectors, we further replace convolution in the task-specific heads of Faster/Mask R-CNN with involution, which could cut down more than half of the computational complexity while retaining the superior or on-par performance. This kind of fully involution-based detectors may stand out especially in cases where computational resource is the major bottleneck. Forth, we pay special attention to the scores of small/medium/large objects and notice that the most compelling performance improvement appears in the measurement of \ap{L}. Our best detection models could surpass the baselines by more than 3\% bounding box AP in this regard, specifically 3.4\%, 4.3\%, and 3.3\% for RetinaNet, Faster R-CNN, and Mask R-CNN. We hypothesize that the success of detecting large-scale objects arises from the design of spread-out and position-aware involution kernels. Besides \ap{L}, the performance gains are consistent under the fine-grained taxonomy of AP evaluation metrics, demonstrated in different columns of Table~\ref{tab:det}.

\vspace{-0.5em}
\subsubsection{Semantic Segmentation}

\begin{table*}[t]
	\centering
	\tablestyle{5pt}{1.0}
	\resizebox{.9\linewidth}{!}{
		\begin{tabular}{c|c|c|c|c|l|l|l|l}
			\fontsize{7pt}{1em}\selectfont Segmentor & \fontsize{7pt}{1em}\selectfont Backbone & \fontsize{7pt}{1em}\selectfont Neck
			& \fontsize{7pt}{1em}\selectfont \#Params (M)
			& \fontsize{7pt}{1em}\selectfont FLOPs (G)
			& \fontsize{7pt}{1em}\selectfont mean IoU (\%) 
			& \fontsize{7pt}{1em}\selectfont ~~~~~~wall 
			& \fontsize{7pt}{1em}\selectfont ~~~~~~truck 
			& \fontsize{7pt}{1em}\selectfont ~~~~~~bus \\
			\shline
			\multirow{3}{*}{Semantic FPN~\cite{Kirillov_2019_CVPR}} & ResNet-50 & convolution
			& 28.5 & 362.8 
			& 74.5 & 39.4 & 58.6 & 72.2 \\
			& RedNet-50 & convolution
			& 18.5 & 293.9 
			& 78.3\cgaphl{+}{3.8} & 52.7\cgaphl{+}{13.3} & 77.3\cgaphl{+}{18.7} & 87.6\cgaphl{+}{15.4}\\
			& RedNet-50 & involution
			& 16.4 & 205.2 
			& 79.2\cgaphl{+}{4.7} & 56.9\cgaphl{+}{17.5} & 82.1\cgaphl{+}{23.5} & 88.5\cgaphl{+}{16.3} \\
		\end{tabular}
	}
	\vskip -0.1in
	\caption{Performance comparison on Cityscapes segmentation based on Semantic FPN. We showcase the mean IoU averaged over all classes, as well as IoUs of the top three classes with the most evident performance amelioration.
	} 
	\label{tab:seg-fpn}
	\vskip -0.2in
\end{table*}

\begin{table}[t]
	\centering
	\tablestyle{5pt}{1.0}
	\resizebox{1.0\linewidth}{!}{
	\begin{tabular}{c|c|c|c|l}
		\fontsize{7pt}{1em}\selectfont Segmentor & \fontsize{7pt}{1em}\selectfont Backbone
		& \fontsize{7pt}{1em}\selectfont \#Params (M)
		& \fontsize{7pt}{1em}\selectfont FLOPs (G)
		& \fontsize{7pt}{1em}\selectfont mIoU (\%) \\
		\shline
		\multirow{2}{*}{UPerNet~\cite{Xiao_2018_ECCV}} & ResNet-50 
		& 66.4 & 1894.5 
		& 78.2 \\
		& RedNet-50 
		& 56.4 & 1825.6 
		& 80.6\cgaphl{+}{2.4} \\
		\hline
		\multirow{3}{*}{Panoptic-DeepLab~\cite{Cheng_2020_CVPR}} 
		& Axial-DeepLab-S~~~ & 12.1 & 220.8 & 80.5 \\
		& Axial-DeepLab-M~~ & 25.9 & 419.6 & 80.3 \\
		& Axial-DeepLab-XL & 173.0 & 2446.8 & 80.6 \\
	\end{tabular}
}
	\vskip -0.1in
	\caption{Performance comparison on Cityscapes segmentation based on UPerNet. The efficiency of UPerNet is greatly boosted by the RedNet backbone, showing competitive performance to Axial-DeepLab-XL with only 32.6\% parameter counts and 75.7\% computational cost.
	} 
	\label{tab:seg-upernet}
	\vskip -0.2in
\end{table}

To further exploit the versatility of involution, we also conduct experiments on the task of semantic image segmentation. We choose the segmentation frameworks of Semantic FPN~\cite{Kirillov_2019_CVPR} and UPerNet~\cite{Xiao_2018_ECCV}, loaded with ImageNet pre-trained backbone weights. We fine-tune these segmentors on the finely-annotated part of the Cityscapes dataset~\cite{Cordts_2016_CVPR}, which contains a split of 2975 and 500 images for training and validation respectively, divided into 19 classes. More training details can be found in the appendix. After training, we perform the evaluation on the validation set under the single-scale mode and adopt the Intersection-over-Union (IoU) as the evaluation metric.

Based on the Semantic FPN framework, we are able to achieve 3.8\% higher mean IoU over all classes, taking advantage of RedNet over ResNet as the backbone. Consequent to further infusing involution into the FPN neck to replace convolution, the gain in mean IoU is elevated to 4.7\% but the parameters and FLOPs are cut down to 57.5\% and 56.6\% of the baseline model accordingly. The detailed comparison results are shown in Table~\ref{tab:seg-fpn}. To take one step further, we investigate the effectiveness of our method on different object classes. Aligned with the discovery in object detection, we notice that the segmentation effects of those objects with a large spatial arrangement are improved by more than 10\%, \eg, wall, truck, and bus, while slight improvements are observed in classes of relatively small objects, \eg, traffic light, traffic sign, person, and bicycle. Once again, the involution operation effectively aids the large object perception by endowing the representation process with dynamic and distant interactions. In addition, we replace the ResNet backbone of UPerNet with RedNet and evaluate the final performance, as displayed in Table~\ref{tab:seg-upernet}. Though not an apple-to-apple comparison using the same segmentor and training strategy, RedNet-based UPerNet appears more efficient than Axial-DeepLab, which is dedicatedly designed for segmentation tasks by converting the original Axial ResNet backbone network.

\subsection{Ablation Analysis}
\label{sec:ablation}

We present several ablation studies designed to understand the contributions of individual components, taking RedNet-50 as an example.

\begin{table*}[t]
	\vspace{-1.em}
	\centering
	\resizebox{\linewidth}{!}{~~~~
		\subfloat[\scriptsize{Accuracy saturates with \textbf{kernel size} increasing.}]{
			\tablestyle{.8pt}{1.05}
			\begin{tabular}{c|c|c|c}
				\fontsize{7pt}{1em}\selectfont Kernel Size
				& \fontsize{7pt}{1em}\selectfont \#Params (M)
				& \fontsize{7pt}{1em}\selectfont FLOPs (G)
				& \fontsize{7pt}{1em}\selectfont Top-1 Acc. (\%) \\
				\shline
				$3 \times 3$ & 14.7 & 2.4 & 76.9 \\
				$5 \times 5$ & 15.1 & 2.5 & 77.4 \\
				\demph{$7 \times 7$} & \demph{15.5} & \demph{2.6} & \demph{77.7} \\
				$9 \times 9$ & 16.2 & 2.7 & 77.8 \\
			\end{tabular}
			\label{tab:kernel-size}
		}~~~~  
		
		\subfloat[\scriptsize{Appropriate \textbf{grouping channels} improves efficiency.}]{
			\tablestyle{.8pt}{1.05}
			\begin{tabular}{c|c|c|c}
				\fontsize{7pt}{1em}\selectfont \#Group Channel
				& \fontsize{7pt}{1em}\selectfont \#Params (M)
				& \fontsize{7pt}{1em}\selectfont FLOPs (G)
				& \fontsize{7pt}{1em}\selectfont Top-1 Acc. (\%) \\
				\shline
				1 & 30.2 & 5.0 & 77.9 \\
				4 & 18.5 & 3.0 & 77.7 \\
				\demph{16} & \demph{15.5} & \demph{2.6} & \demph{77.7} \\
				$C$ & 14.6 & 2.4 & 76.5 \\
			\end{tabular}
			\label{tab:group-channel}
		}~~~~  

		\subfloat[\scriptsize{Introducing the \textbf{bottleneck structure} reduces complexity.}]{
			\tablestyle{.8pt}{1.05}
			\begin{tabular}{c|c|c|c}
				\fontsize{7pt}{1em}\selectfont Function Form
				& \fontsize{7pt}{1em}\selectfont \#params (M)
				& \fontsize{7pt}{1em}\selectfont FLOPs (G)
				& \fontsize{7pt}{1em}\selectfont Top-1 Acc. (\%) \\
				\shline
				$\mathbf{W}$ & 18.1 & 3.0 & 77.8 \\
				$\mathbf{W}_1 \sigma \mathbf{W}_0, r=1$~~ & 19.4 & 3.2 & 77.8 \\
				\demph{$\mathbf{W}_1 \sigma \mathbf{W}_0, r=4$}~~ & \demph{15.5} & \demph{2.6} & \demph{77.7} \\
				$\mathbf{W}_1 \sigma \mathbf{W}_0, r=16$ & 14.6 & 2.4 & 77.4 \\
			\end{tabular}
			\label{tab:function-type}
		}~~~~  
	}  
	\vskip -0.1in
	\caption{We examine the role of different components in the design of involution, including kernel size, group channels, and the form of kernel generation function. In gray are entries with the {\color{Gray}{default setting}} as our main experiments. When we adjust one hyper-parameter for ablation, we leave the others as the default setting. The final performance is not sensitive to each hyper-parameter configuration.
	}
	\label{tab:ablation}
	\vskip -0.2in
\end{table*}

\vspace{-1.5em}
\paragraph{Stem}

First of all, we isolate the impact of involution on the network stem. Following the practice of recent self-attention based architectures~\cite{Zhao_2020_CVPR,Wang_2020_ECCV}, the network stem is decomposed into three consecutive operations to save memory cost. In accordance with our practice of integrating involution into the trunk, we place $3\times 3$ involution at the bottleneck position of the stem. This act improves the accuracy from 77.7\% to 78.4\% with marginal cost, leading to our default setting of RedNet in the main experiments. 

Otherwise explicitly mentioned, we use RedNet-50 with \emph{$7 \times 7$ convolution stem} for the following ablation analysis.

\vspace{-1.5em}
\paragraph{Kernel Size}

In the spatial dimension, we probe the effect of kernel size. Steady improvement is observed in Table~\ref{tab:kernel-size} when increasing the spatial extent up to $7 \times 7$ with negligible computational overheads. The improvement somewhat plateaus when further expanding the spatial extent, which is possibly relevant to the feature resolution in the network. This set of controlled experiments shows the superiority of harnessing large involution kernels over compact and static convolution, while avoiding to introduce prohibitive memory and computational cost.

\vspace{-1.5em}
\paragraph{Group Channel}

In the channel dimension, we assess the feasibility of sharing an involution kernel. As can be seen in Table~\ref{tab:group-channel}, sharing a kernel per 16 channels halves the parameters and computational cost compared to the non-shared one, only sacrificing 0.2\% accuracy. However, sharing a single kernel across all the $C$ channels obviously underperforms in accuracy. Considering the channel redundancy of involution kernels, as long as setting the channels shared in a group to an acceptable range, the channel-agnostic behavior will not only reserve the performance, but also reduce the parameter count and computational cost. This will also permit a larger kernel size under the same budget.

\begin{figure}[t]
	\begin{center}
		\begin{minipage}{.937\linewidth}
			\includegraphics[width=.19\linewidth]{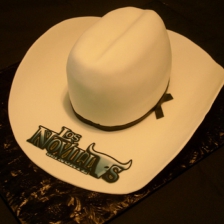}~
			\includegraphics[width=.19\linewidth]{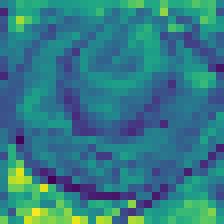}
			\includegraphics[width=.19\linewidth]{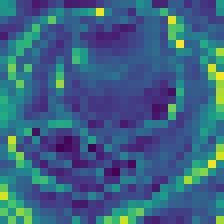}
			\includegraphics[width=.19\linewidth]{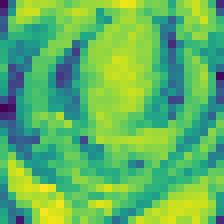}
			\includegraphics[width=.19\linewidth]{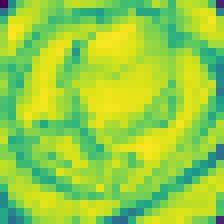}
			\includegraphics[width=.19\linewidth]{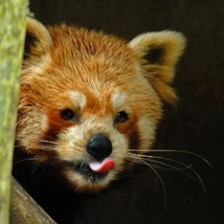}~
			\includegraphics[width=.19\linewidth]{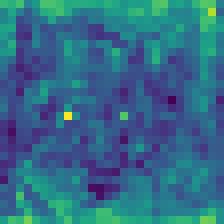}
			\includegraphics[width=.19\linewidth]{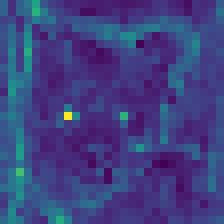}
			\includegraphics[width=.19\linewidth]{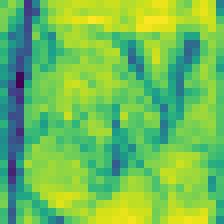}
			\includegraphics[width=.19\linewidth]{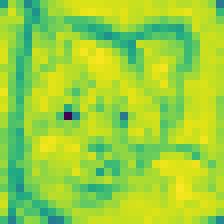}
			\includegraphics[width=.19\linewidth]{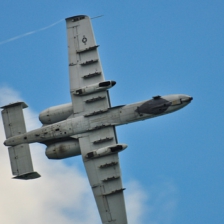}~
			\includegraphics[width=.19\linewidth]{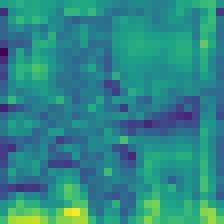}
			\includegraphics[width=.19\linewidth]{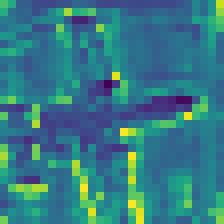}
			\includegraphics[width=.19\linewidth]{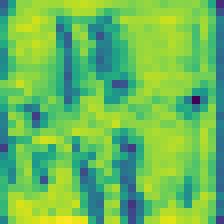}
			\includegraphics[width=.19\linewidth]{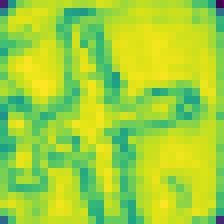}
			\includegraphics[width=.19\linewidth]{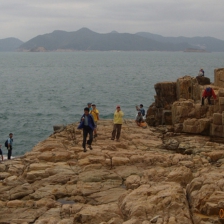}~
			\includegraphics[width=.19\linewidth]{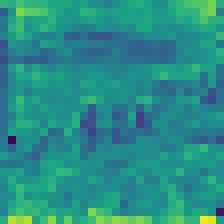}
			\includegraphics[width=.19\linewidth]{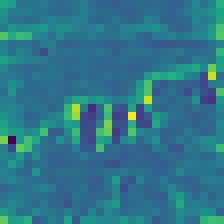}
			\includegraphics[width=.19\linewidth]{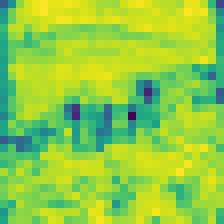}
			\includegraphics[width=.19\linewidth]{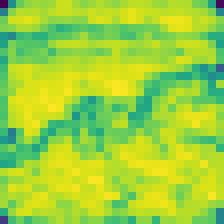}
		\end{minipage}
		\begin{minipage}{.053\linewidth}
			\includegraphics[width=\linewidth]{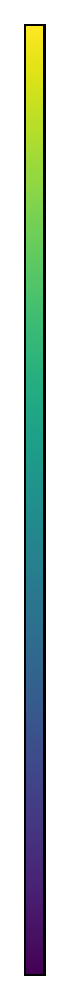}
		\end{minipage}
	\end{center}
	\vskip -0.25in
	\caption{The heat maps in each row interpret the generated kernels for an image instance from the ImageNet validation set, drawn from four different classes, including cowboy hat, lesser panda, warplane, and cliff (from top to bottom).}
	\label{fig:visualization}
	\vskip -0.2in
\end{figure}

\vspace{-1.5em}
\paragraph{Kernel Generation Function}

Next, we validate the utility of bottleneck architecture for the kernel generation process in Table~\ref{tab:function-type}. Adopting a single linear transform $\mathbf{W}$ or two transforms without bottleneck ($r=1$) as the kernel generation function incurs more parameters and FLOPs but only performs marginally better, compared to the default setting ($r=4$). Moreover, inferior performance could be ascribed to aggressive channel reduction ($r=16$). 

Further attaching activation functions such as softmax, sigmoid to the kernel generation function, would constrain the kernel values, thus restrict its expressive capability, and ends up hindering the performance by over 1\%. So we opt not to insert any additional functions at the output end of the kernel generation function, allowing the generated kernel to span the entire subspace of $K \times K$ matrices.

\subsection{Visualization}

For dissecting the learned involution kernels, we take the sum of $K \times K$ values from each involution kernel as its representative value. All the representatives at different spatial locations frame the corresponding heat map. Some selected heat maps are plotted in Figure~\ref{fig:visualization}, where the columns following the original image indicate different involution kernels in the last block of the third stage (conv3\_4 following the naming convention of~\cite{He_2016_CVPR}), separated by groups. On the one hand, involution kernels automatically attend to crucial parts of objects in the spatial range for correct image recognition. On the other hand, in a single involution layer, different kernels from different groups focus on varying semantic concepts of the original image, by highlighting peripheral parts, sharp edges or corners, smoother regions, outline of the foreground and background objects, respectively (from left to right in each row).

\section{Conclusion and Prospect}

In this work, we present involution, an effective and efficient operator for visual representation learning, reversing the design principles of convolution and generalizing the formulation of self-attention. Thanks to the medium of involution, we are able to disclose the underlying relationship between self-attention and convolution and empirically ascertain the essential driving force of self-attention for its recent progress in vision. Our proposed involution is benchmarked on several standard vision benchmarks, consistently delivering enhanced performance at reduced cost compared to convolution-based counterparts and self-attention based models. Furthermore, careful ablation analysis helps us better understand that such performance enhancement is rooted in the core contributions of involution, from the efficacy of spatial modeling to the efficiency of architecture design.

We believe that this work could foster future research enthusiasm on simple yet effective visual primitives beyond convolution, which is expected to make inroads into fields of neural architecture engineering where uniform and local spatial modeling has prestigiously dominated.

\appendix

\section{Implementation Details}

\subsection{Image Classification}

In accordance with Stand-Alone Self-Attention~\cite{NIPS2019_8302} and Axial Attention~\cite{Wang_2020_ECCV}, we train all these models for 130 epochs utilizing the Stochastic Gradient Descent (SGD) optimizer with the momentum of 0.9 and the weight decay of 0.0001. The learning rate initiates from 0.8 and gradually approaches zero following a half-cosine function shaped schedule. The mini-batch size per GPU is set to 32 and the training procedure is conducted on 64 GPU devices in total. The label smoothing regularization technique is applied with the coefficient of 0.1.

\subsection{Object Detection and Instance Segmentation}

Following the widely-adopted pipeline, the input images are resized to keep their shorter/longer side as 800/1333 pixels prior to being fed into the networks. The training procedure lasts for 12 epochs, using the Stochastic Gradient Descent (SGD) optimizer with the momentum of 0.9 and weight decay of 0.0001. The initial learning rate is set to 0.02 for Faster/Mask R-CNN and 0.01 for RetinaNet with a linear warm-up period of 500 iterations, divided by 10 in the 8\textsuperscript{th} and 11\textsuperscript{st} epoch. When necessary, we moderately extend the warm-up period and apply gradient clipping for the sake of convergence stability. The detectors are trained on 8 Tesla V100 GPUs with 2 samples per GPU.

\subsection{Semantic Segmentation}

The urban scene images with a high resolution of $1024 \times 2048$ are randomly resized, with their aspect ratios kept in the range from 0.5 to 2.0, from which the input image patches with the size of $512 \times 1024$ are randomly cropped, then undergo random horizontal flipping and a sequence of photometric distortions as the data augmentation. We adopt the training schedule of 80k iterations, and apply the Stochastic Gradient Descent (SGD) optimizer with the momentum of 0.9 and weight decay of 0.0005. The learning rate starts from 0.01 and anneals following the conventional ``poly'' policy, which indicates the initial learning rate is multiplied by $(1 - \frac{iter}{total\_iter})^{0.9}$ in each iteration. The segmentation networks are trained on 4 Tesla V100 GPUs with 2 samples per GPU. We apply synchronized Batch Normalization~\cite{Peng_2018_CVPR} for more stable estimation of the batch statistics.

\section{Comparison to State-of-the-art on COCO}

\begin{table}[t]
	\centering
	\tablestyle{5pt}{1.0}
	\resizebox{\linewidth}{!}{
		\begin{tabular}{c|ccc|ccc}
			\fontsize{7pt}{1em}\selectfont Method 
			& \fontsize{7pt}{1em}\selectfont \apbbox{} 
			& \fontsize{7pt}{1em}\selectfont \apbbox{50} 
			& \fontsize{7pt}{1em}\selectfont \apbbox{75} 
			& \fontsize{7pt}{1em}\selectfont \apmask{} 
			& \fontsize{7pt}{1em}\selectfont \apmask{50} 
			& \fontsize{7pt}{1em}\selectfont \apmask{75} \\
			\shline
			baseline & 38.4 & 59.2 & 41.9 & 35.1 & 56.3 & 37.3 \\
			+ NL~\cite{Wang_2018_CVPR} & 39.0 & 61.1 & 41.9 &  35.5 & 58.0 & 37.4 \\
			+ RCCA~\cite{Huang_2019_ICCV} & 39.3 & - & - & 36.1 & - & - \\
			+ GC @C5~\cite{Cao_2019_ICCV} & 38.7 & 61.1 & 41.7 & 35.2 & 57.4 & 37.4 \\
			+ DCN @C5~\cite{Zhu_2019_CVPR} & 39.9 & - & - & 34.9 & - & - \\
			+ DGMN @C5~\cite{Zhang_2020_CVPR} & 40.2 & 62.0 & 43.4 & 36.0 & 58.3 & 38.2 \\
			\shline
			\textbf{ours} & \textbf{40.8} & \textbf{62.3} & \textbf{44.3} & \textbf{36.4} & \textbf{59.0} & \textbf{38.5} \\
		\end{tabular}
	}
	\vskip -0.1in
	\caption{Quantitative comparison on the COCO 2017 validation set. Our model could outstrip the previous methods with attention or dynamic add-on, using reduced parameters and computational cost. C5 indicates inserting the considered components at all the $3 \times 3$ convolution layers of the last stage (conv5\_x) in ResNet-50.
	} 
	\label{tab:det-supp}
	\vskip -0.15in
\end{table}

For both object detection and instance segmentation on COCO, we compare our involution-based Mask R-CNN~\cite{He_2017_ICCV} with the RedNet-50 backbone against other celebrated architectures with ResNet-50 in Table~\ref{tab:det-supp}. Our approach performs substantially better than convolution-based Mask R-CNN equipped with self-attention blocks, like NLNet~\cite{Wang_2018_CVPR}, CCNet~\cite{Huang_2019_ICCV}, and GCNet~\cite{Cao_2019_ICCV}. Additionally, our method outperforms those of embedding dynamic mechanism into the networks, including Deformable ConvNets (DCN)~\cite{Zhu_2019_CVPR} and Dynamic Graph Message passing Networks (DGMN)~\cite{Zhang_2020_CVPR}. Note that all these referred approaches introduce extra parameters and FLOPs to the vanilla Mask R-CNN by appending complementary modules while our proposed involution operator even reduces the complexity of baseline by substituting convolution.

\section{Visualization of Segmentation}

\begin{figure*}[t]
	\begin{center}
		\includegraphics[width=.33\linewidth]{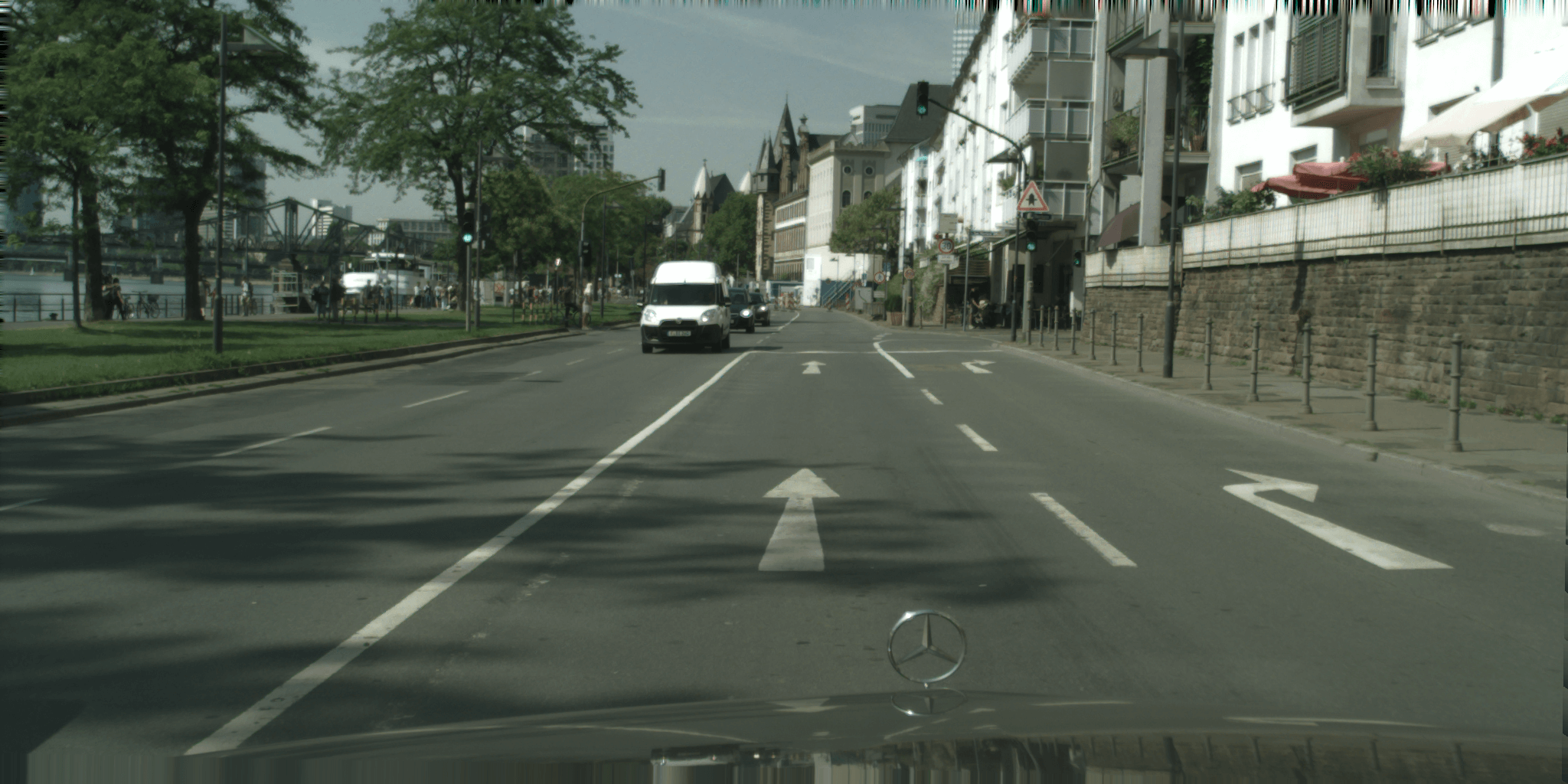}
		\includegraphics[width=.33\linewidth]{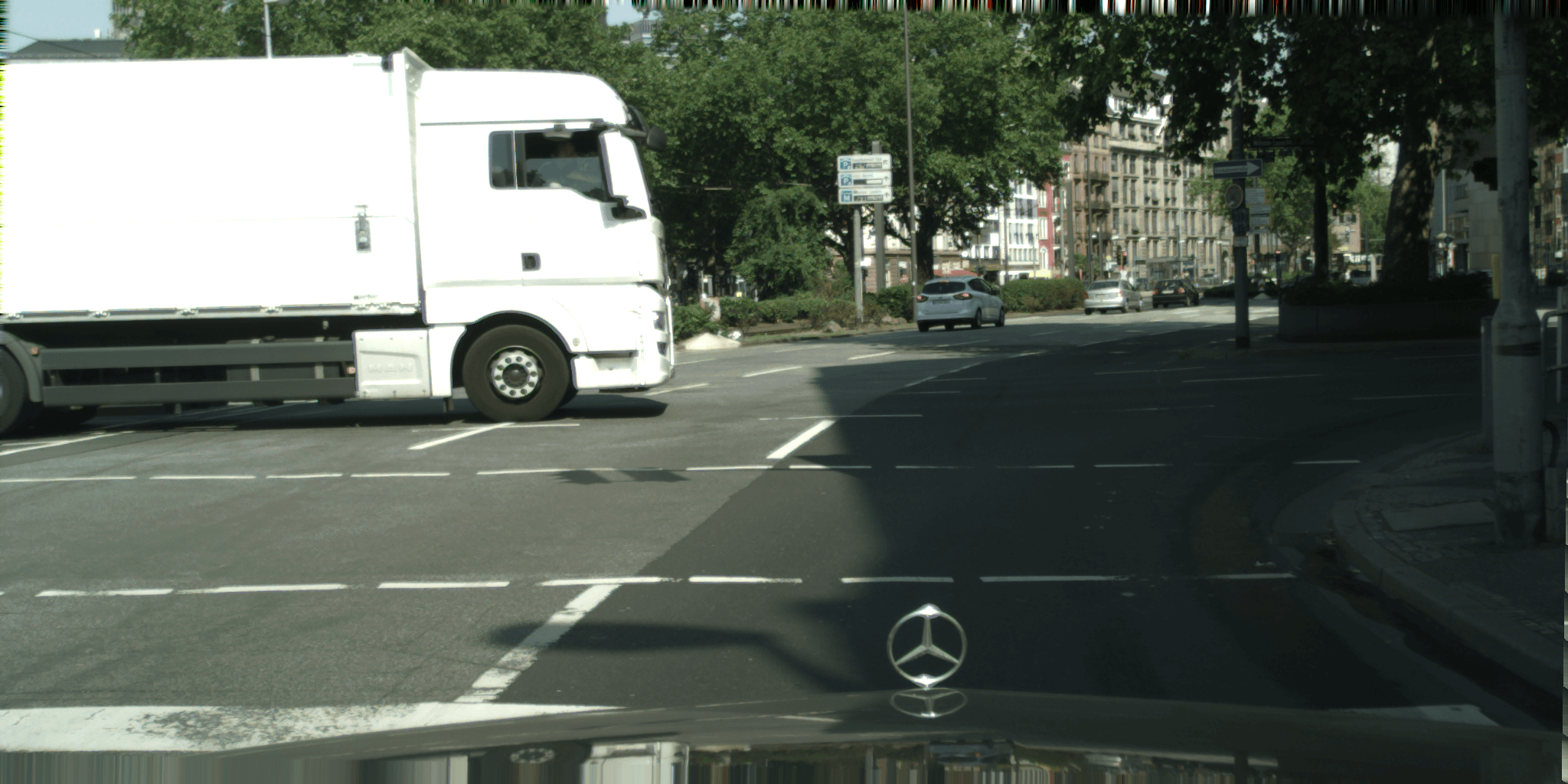}
		\includegraphics[width=.33\linewidth]{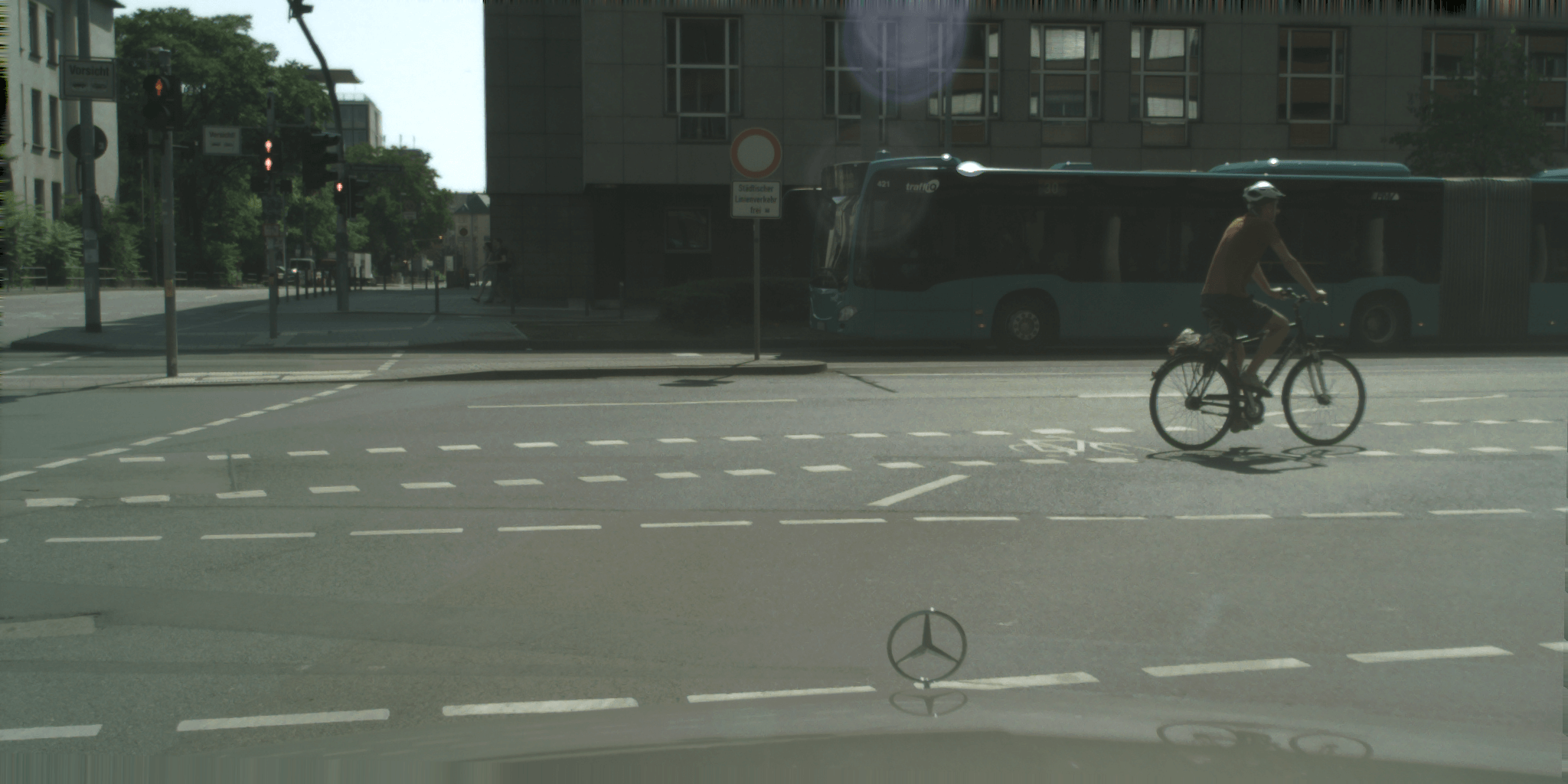}
		\includegraphics[width=.33\linewidth]{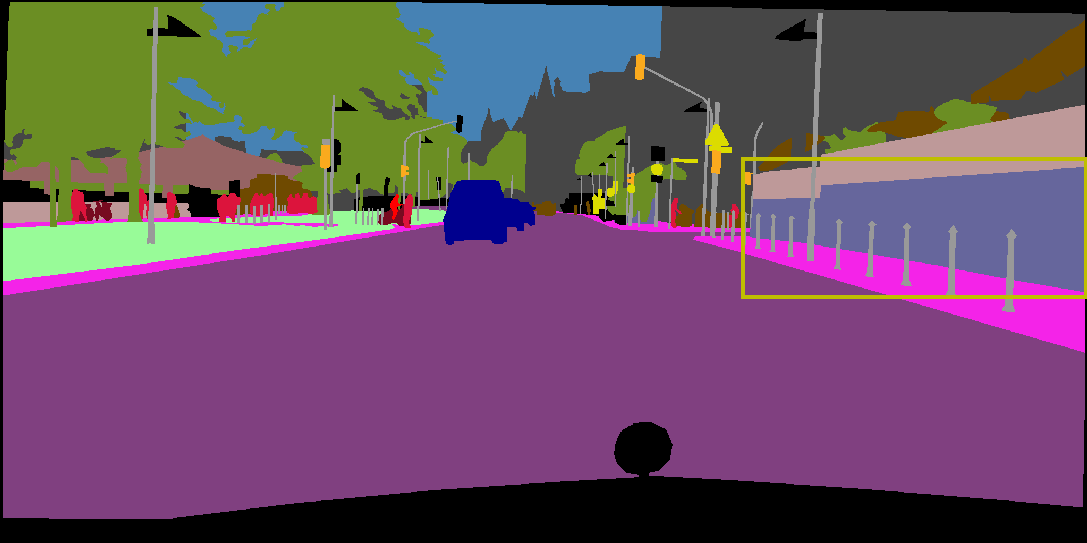}
		\includegraphics[width=.33\linewidth]{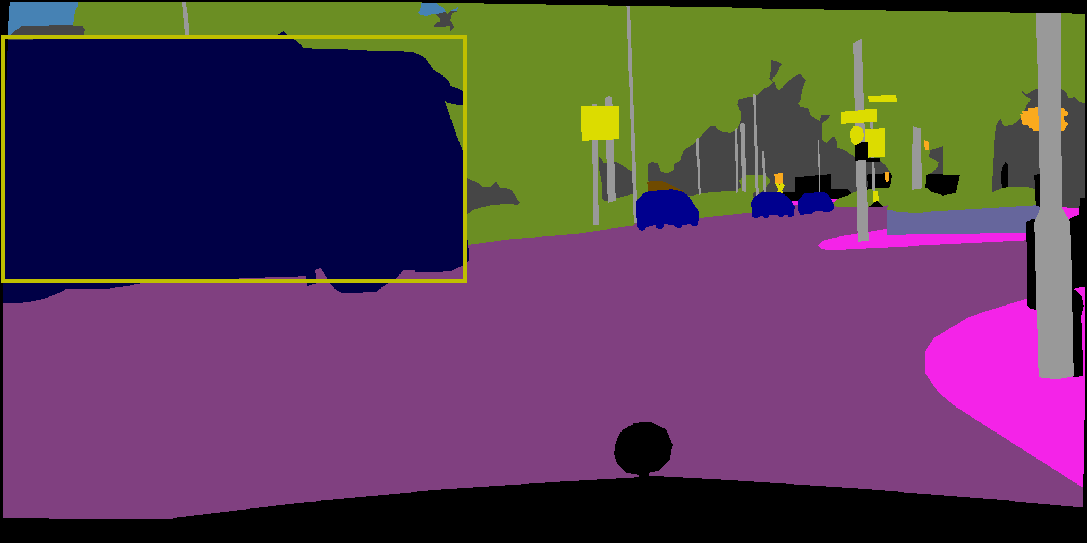}
		\includegraphics[width=.33\linewidth]{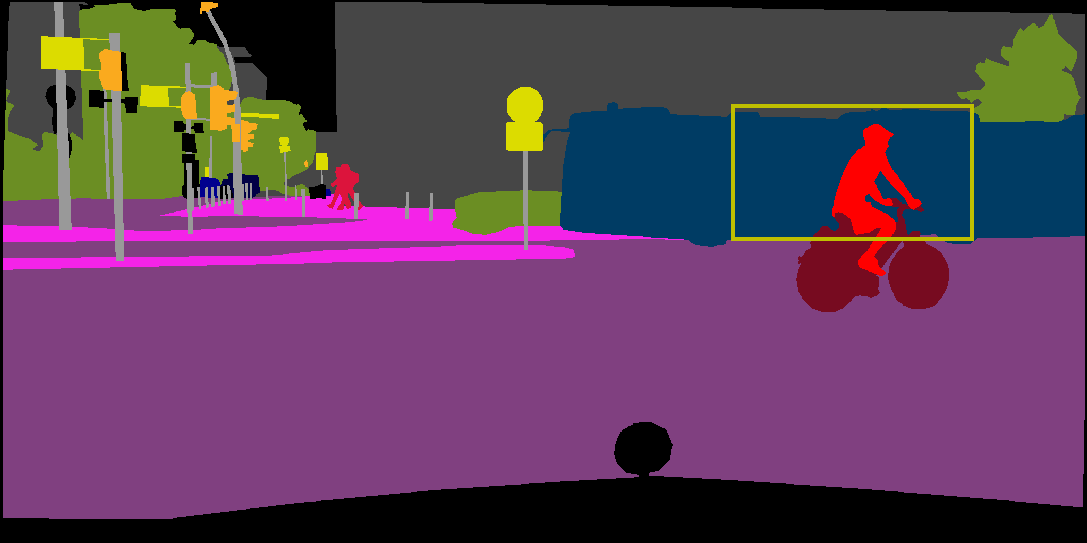}
		\includegraphics[width=.33\linewidth]{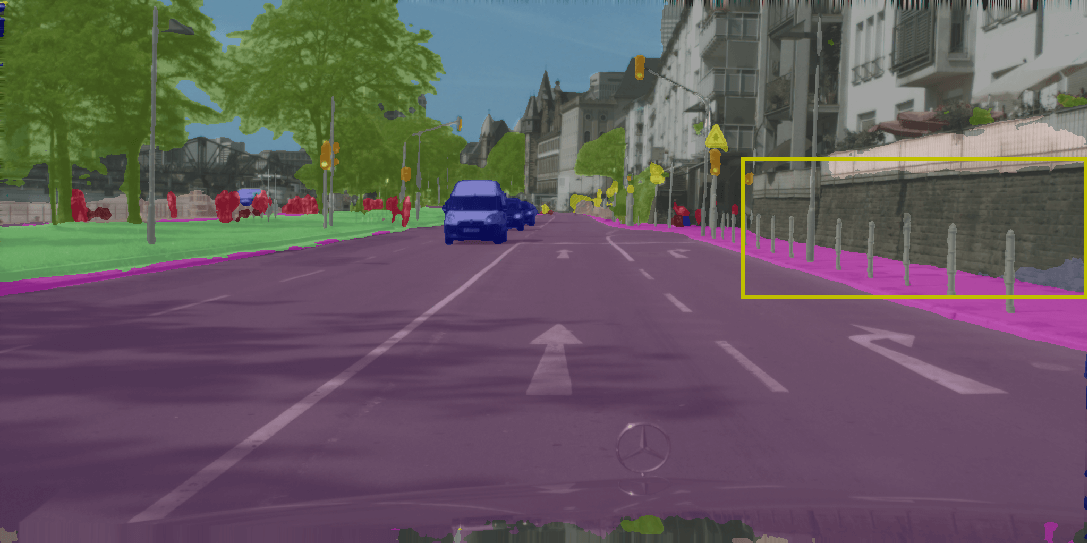}
		\includegraphics[width=.33\linewidth]{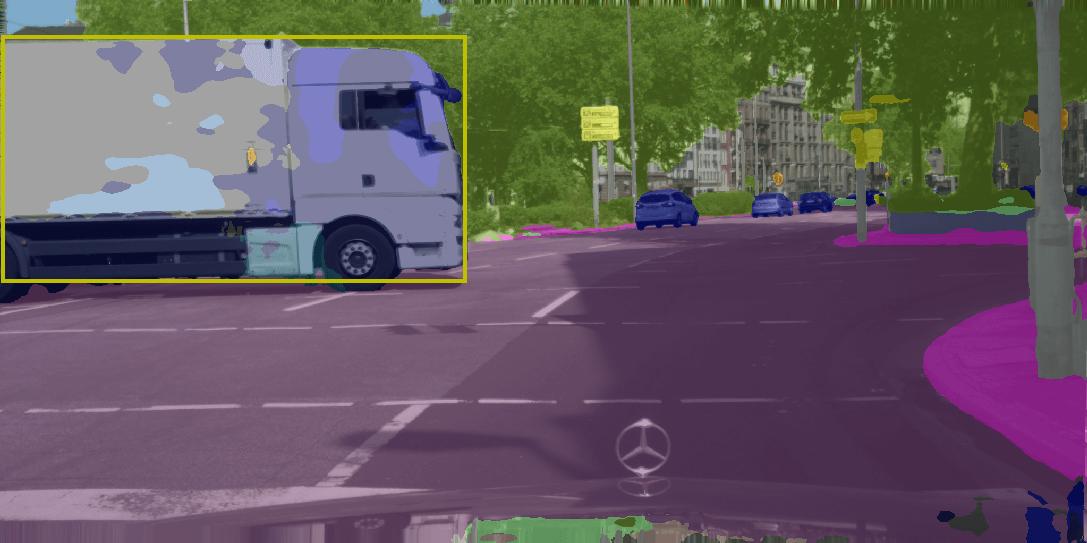}
		\includegraphics[width=.33\linewidth]{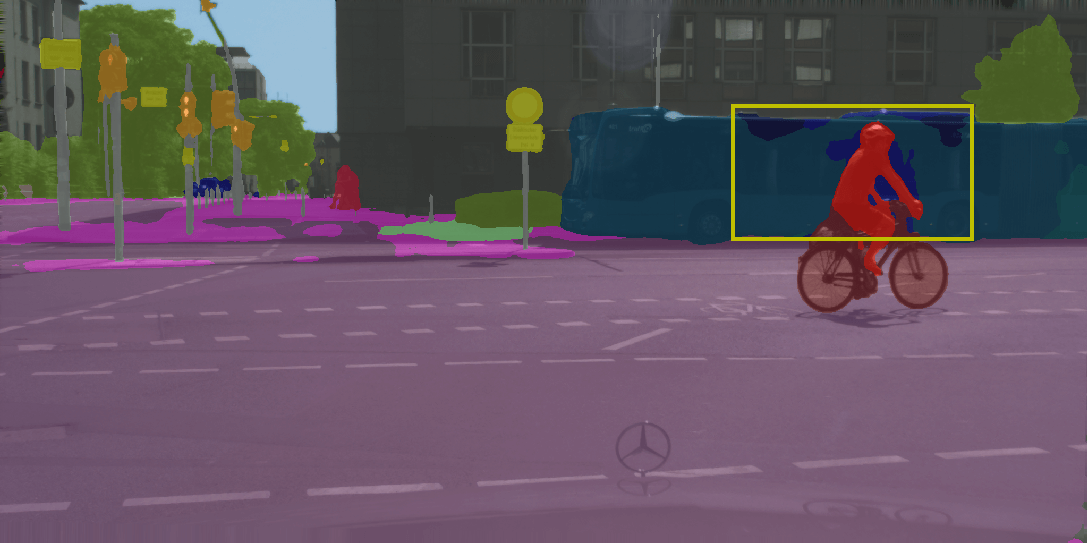}
		\includegraphics[width=.33\linewidth]{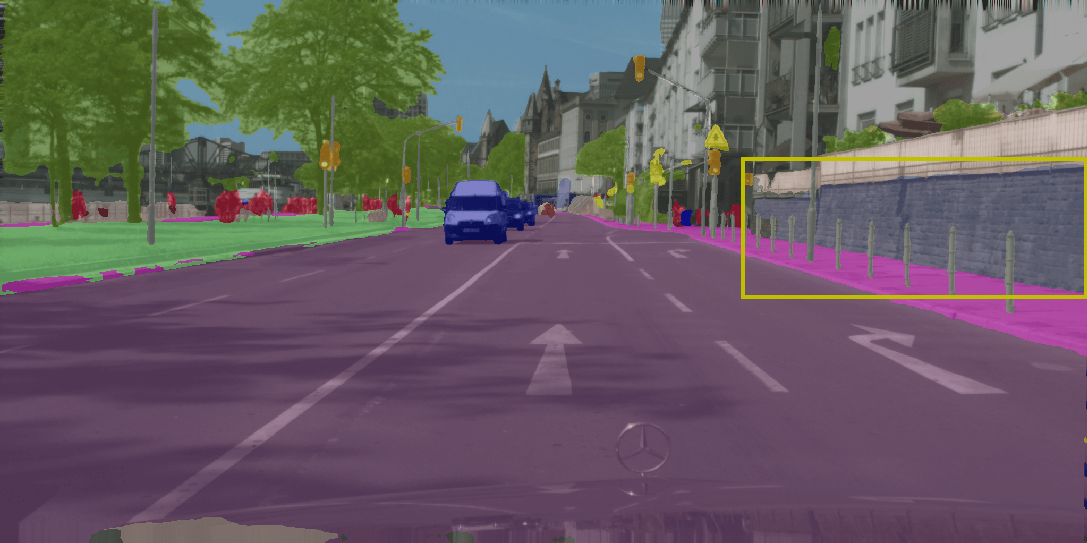}
		\includegraphics[width=.33\linewidth]{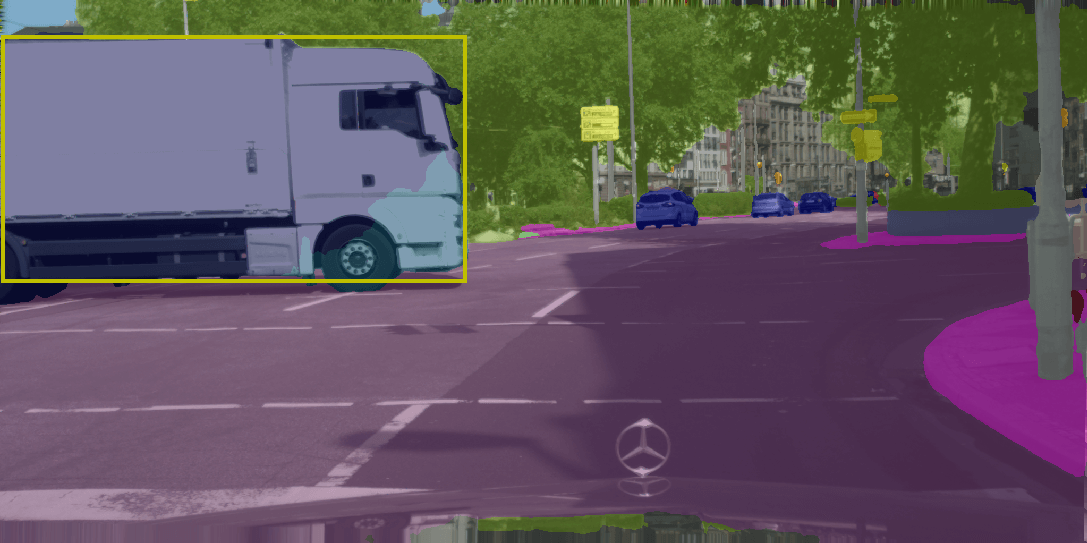}
		\includegraphics[width=.33\linewidth]{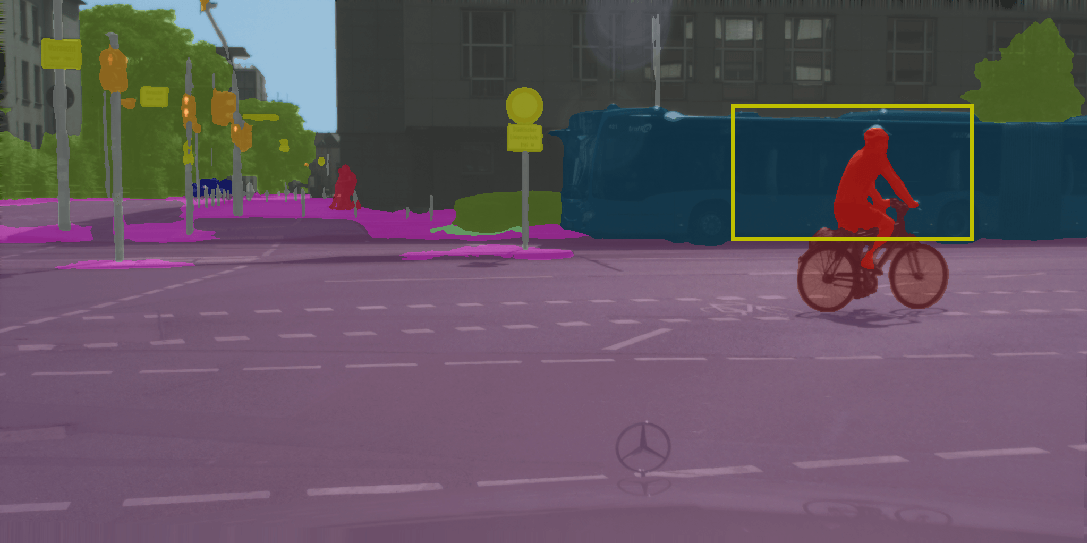}
	\end{center}
	\caption{Qualitative comparison of segmentation results on the Cityscapes validation set. Each column represents an image example of urban scene. The first and second row show the original image and ground truth. The last two rows demonstrate the prediction results of baseline and our method, respectively. Highlighted in the yellow boxes are their apparent differences.}
	\label{fig:seg}
\end{figure*}

Based on the semantic FPN~\cite{Kirillov_2019_CVPR} framework, we provide some prediction results on the Cityscapes validation set in Figure~\ref{fig:seg}. Without the help of involution, pixels of large objects are usually mistaken as other objects with high similarity. For instance, the wall in the first image example are mostly confused with building by the  convolution-based FPN. Some pixels of the bus in the third image example are misclassified as truck or car, distracted by the occlusion of the cyclist. In contrast, our involution-based FPN dissolves these ambiguities by dynamically reasoning in an enlarged spatial range. Also, better consistency of inner pixels of an object is observed in the segmentation results of our method, reaping the benefits of involution.

\section{Discussion}

The topological connectivity~\cite{He_2016_CVPR,Huang_2017_CVPR,Xie_2019_ICCV,Yang_2018_CVPR} and hyper-parameter configurations~\cite{Guo_2020_ECCV,Radosavovic_2020_CVPR,pmlr-v97-tan19a} of convolutional neural networks have undergone rapid evolution, but developing brand new operators attracts little attention for crafting innovative architectures. In this work, we expect to bridge this regret via disassembling the elements of convolution and reassembling them into a more effective and efficient involution. In the meanwhile, one of the current front edges of neural architecture engineering is automatically searching the network structures~\cite{cai2018proxylessnas,liu2018darts,pmlr-v80-pham18a,Zoph2017Neural,Zoph_2018_CVPR}.  Our invention can also fill the pool of search space for most existing Neural Architecture Search (NAS) strategies. In the near future, we are looking forward to discovering more effective involution-equipped neural networks with the help of NAS.

{\small
\bibliographystyle{ieee_fullname}
\bibliography{egbib}
}

\end{document}